\documentclass[10pt,twocolumn,letterpaper]{article}

\usepackage{wacv}              

\usepackage{graphicx}
\usepackage{amsmath}
\usepackage{amssymb}
\usepackage{booktabs}
\usepackage{multirow}
\usepackage{tabularray}
\usepackage{siunitx}

\usepackage[pagebackref,breaklinks,colorlinks]{hyperref}
\usepackage[ruled,linesnumbered,noend]{algorithm2e}

\newcommand{\whitespace}{\textrm{$\ $}}

\newcommand{\pointCloud}{\textrm{$P$}}

\newcommand{\mlModel}{\textrm{$M$}}

\newcommand{\binX}{\textrm{$m$}}
\newcommand{\binY}{\textrm{$n$}}

\usepackage[capitalize]{cleveref}
\crefname{section}{Sec.}{Secs.}
\Crefname{section}{Section}{Sections}
\Crefname{table}{Table}{Tables}
\crefname{table}{Tab.}{Tabs.}

\def\wacvPaperID{2736} 
\def\confName{WACV}
\def\confYear{2025}

\begin{document}

\title{On-the-Fly Object-aware Representative Point Selection in Point Cloud}


\author{Xiaoyu Zhang\\
RMIT University\\
{\tt\small s3500791@student.rmit.edu.au}
\and
Ziwei Wang\\
CSIRO\\
{\tt\small ziwei.wang@data61.csiro.au}
\and
Hai Dong\\
RMIT University\\
{\tt\small hai.dong@rmit.edu.au}
\and
Zhifeng Bao\\
RMIT University\\
{\tt\small zhifeng.bao@rmit.edu.au}
\and
Jiajun Liu\\
CSIRO\\
{\tt\small jiajun.liu@csiro.au}
}

\maketitle

\begin{abstract}
Point clouds are essential for object modeling and play a critical role in assisting driving tasks for autonomous vehicles (AVs). However, the significant volume of data generated by AVs creates challenges for storage, bandwidth, and processing cost. To tackle these challenges, we propose a representative point selection framework for point cloud downsampling, which preserves critical object-related information while effectively filtering out irrelevant background points. Our method involves two steps:
(1) Object Presence Detection, where we introduce an unsupervised density peak-based classifier and a supervised Naïve Bayes classifier to handle diverse scenarios, and (2) Sampling Budget Allocation, where we propose a strategy that selects object-relevant points while maintaining a high retention rate of object information.
Extensive experiments on the KITTI and nuScenes datasets demonstrate that our method consistently outperforms state-of-the-art baselines in both efficiency and effectiveness across varying sampling rates. As a model-agnostic solution, our approach integrates seamlessly with diverse downstream models, making it a valuable and scalable addition to the 3D point cloud downsampling toolkit for AV applications.
\end{abstract}

\section{Introduction}
\label{sec:intro}

In the domain of autonomous vehicle (AV) applications, point clouds play a crucial role in modeling the environments. Real-time analysis of these point clouds enables vehicles to perceive their surroundings and supports safe driving by facilitating key perception tasks such as object detection, semantic segmentation, and motion prediction\cite{cui2021deep}. 
However, processing the point cloud data collected by AVs poses a significant challenge due to two primary reasons: the size of individual point clouds being large, and the high frequency at which they are generated. 
Specifically, a typical point cloud representing a single scene comprises 20,000 to 30,000 points~\cite{li2020deep}, and these are captured at high frequencies, such as 10Hz for datasets like KITTI~\cite{kitti2019iccv} and 2Hz for nuScenes~\cite{caesar2020nuscenes}.

\noindent\textbf{Motivation.} The substantial data volume generated by AVs would impose burdens on infrastructure, including the storage, network, and computation resources: \textbf{1)} Storage and Network Constraints. The AV data sometimes would be collected and uploaded to a remote cloud for further processing, e.g., to fine-tune the existing models. AVs generate approximately 5 TB of point cloud data per hour, a volume that tests the limits of current 5G network capabilities, which can optimally handle up to 4.5 TB per hour~\cite{kazhamiaka2021challenges}. This unmatching necessitates reducing data size before transmission to remote servers for further processing. \textbf{2)} Hardware Limitations in Data Processing. The scale of point cloud data imposes high demands on computational resources, particularly during the training of machine learning models. High-performance GPUs, although effective, represent a significant expense~\cite{joshi2023arima}. Thus, we need to find an alternative approach that reduces the data size to alleviate the need for hardware upgrades, thereby enabling more cost-efficient data processing.

\noindent\textbf{Limitations and challenges.} Existing point cloud down-sampling strategies mainly have three limitations:  
\textbf{1)} Scalability: Deep learning-based methods, such as those proposed in\cite{dovrat2019learning, lang2020samplenet}, are primarily designed for single-object point clouds and face significant challenges when extended to large-scale outdoor point clouds. This limitation arises from the inherent complexity and uneven point distribution characteristic of outdoor environments.
\textbf{2)} Generalization capability. There are studies that integrate the point cloud subsampling as a module of the object detection model. Such a module is trained end-to-end~\cite{zhang2022not, hu2020randla}. The sampling result is downstream model specific and cannot generalize to others models. \textbf{3)} Processing efficiency. Deep learning-based downsampling methods rely on feature extraction, which often becomes a bottleneck for computational efficiency. On the other hand, traditional non-learning-based solutions, such as Farthest Point Sampling (FPS), suffer from high computational complexity of $\theta(n^2)$, making it unsuitable for real-time applications.

\begin{figure}[t]
    \centering
    \includegraphics[width=0.4\textwidth]{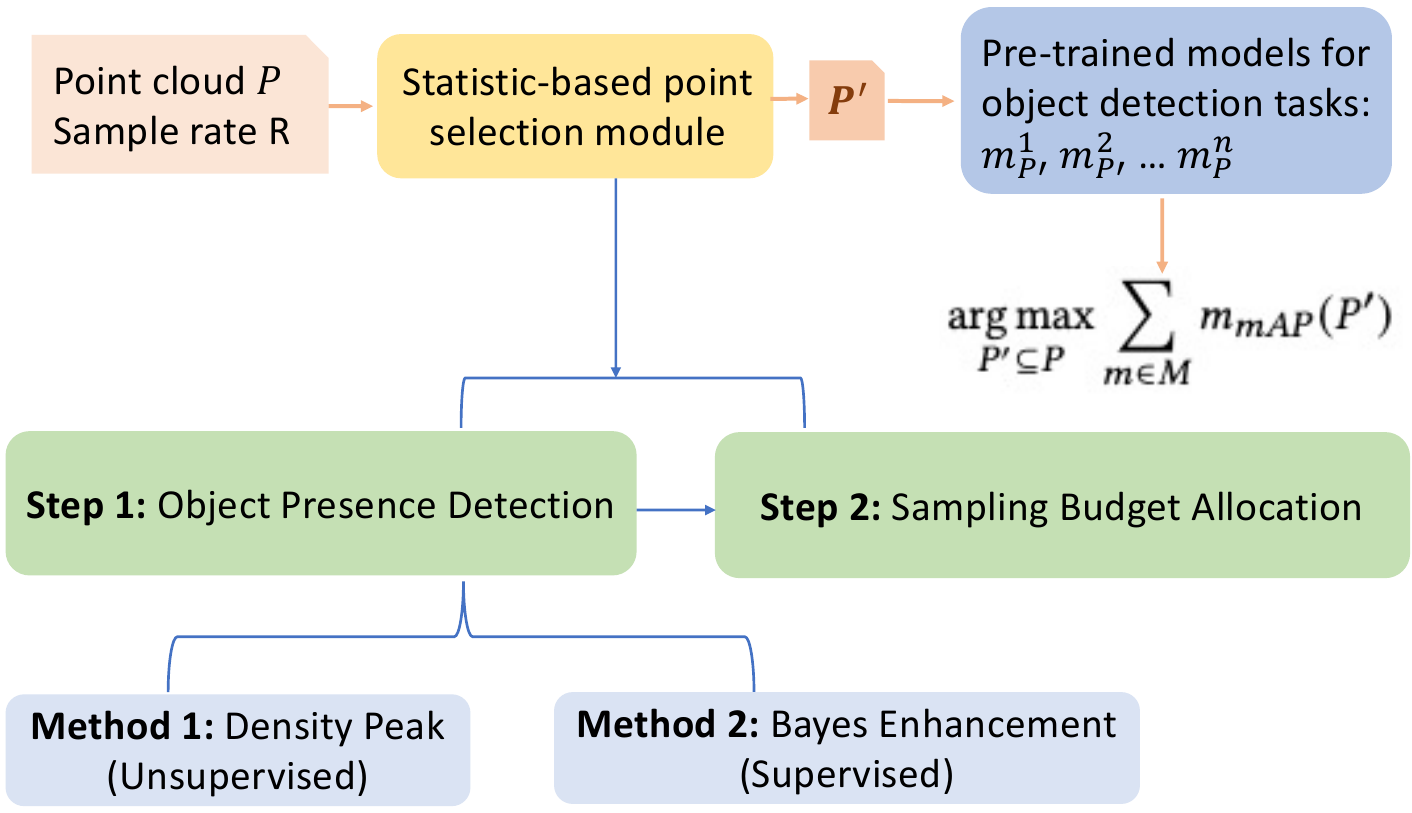}
    \caption{Overview}
    \label{fig: overview}
\end{figure}

\noindent\textbf{Problem Formulation.} 
We propose an object-aware representative point selection approach, which aims to best describe the main content of the point cloud by selecting a subset of points. The subset of points would ideally generalize to various detection models. As demonstrated in Figure \ref{fig: overview}, we achieve the goal by subsampling. First, we classify points in a point cloud into two categories: important object points and insignificant background points. To achieve this, we propose an unsupervised method that utilizes density peaks as classification features. Furthermore, we propose a supervised Na\"ive Bayes-enhanced classifier to improve the classification performance. Second, we propose an imbalanced sample budget allocation with different sampling granularity for the two categories of points, which achieves a reasonable trade-off between effectiveness and efficiency. 

To summarise, we make the following contributions:

\noindent \textbf{1)} By identifying the challenge posed by the significant data size of point cloud from autonomous vehicles, we define our point selection problem, which aims at finding a representative point set in a point cloud and thereby decrease the overall data size (Section \ref{sec: definition}). 

\noindent \textbf{2)} We propose a two-step sampling method, including classifying object points and background points by detecting the presence of the object and allocating different sampling budgets to object points and background points (Section \ref{sec: method}).

\noindent \textbf{3)} To detect the presence of an object, we propose an unsupervised density peak classifier (in Section \ref{sec:method_peak}), and a supervised method with Na\"ive Bayes enhancement (in Section \ref{sec:method_bayes}) to further improve the effectiveness. The two selection methods also provide options to suit supervised and unsupervised application scenarios.

\noindent \textbf{4)} The experimental results demonstrate that our methods exhibit significant advantages in both effectiveness and efficiency, achieving a tenfold increase in speed. Also, our approach is compatible with various downstream models.

\section{Problem Definition and Related Work}
In this section, we provide a formal problem definition and review its related work.

\subsection{Problem Definition}
\label{sec: definition}

Given a point cloud \pointCloud, where each point in \pointCloud \whitespace is formed as a $d$-dimension feature vector $p_i \in \mathbb{R}^{d}$, a set of pre-trained machine learning models \mlModel \whitespace that perform object detection as the downstream task, and a sample rate $r$, we aim to find a method $F(P, r)$ to obtain a subset $P^* \subseteq P$ (where $|P^*| = |P| \times r$) as the representative point set. The objective is to maximize the evaluation result of the object detection task, measured using mean Average Precision (mAP),
by selecting $P^*$ for $m\in\mlModel$: 

\begin{equation}
    P^* = \mathop{\arg\max}_{P' \subseteq P}\sum_{m\in\mlModel} m_{mAP}(P')
\end{equation}

\begin{figure*}[!t]
    \centering
    \begin{subfigure}{0.18\linewidth}
        \centering
        \includegraphics[width=\textwidth]{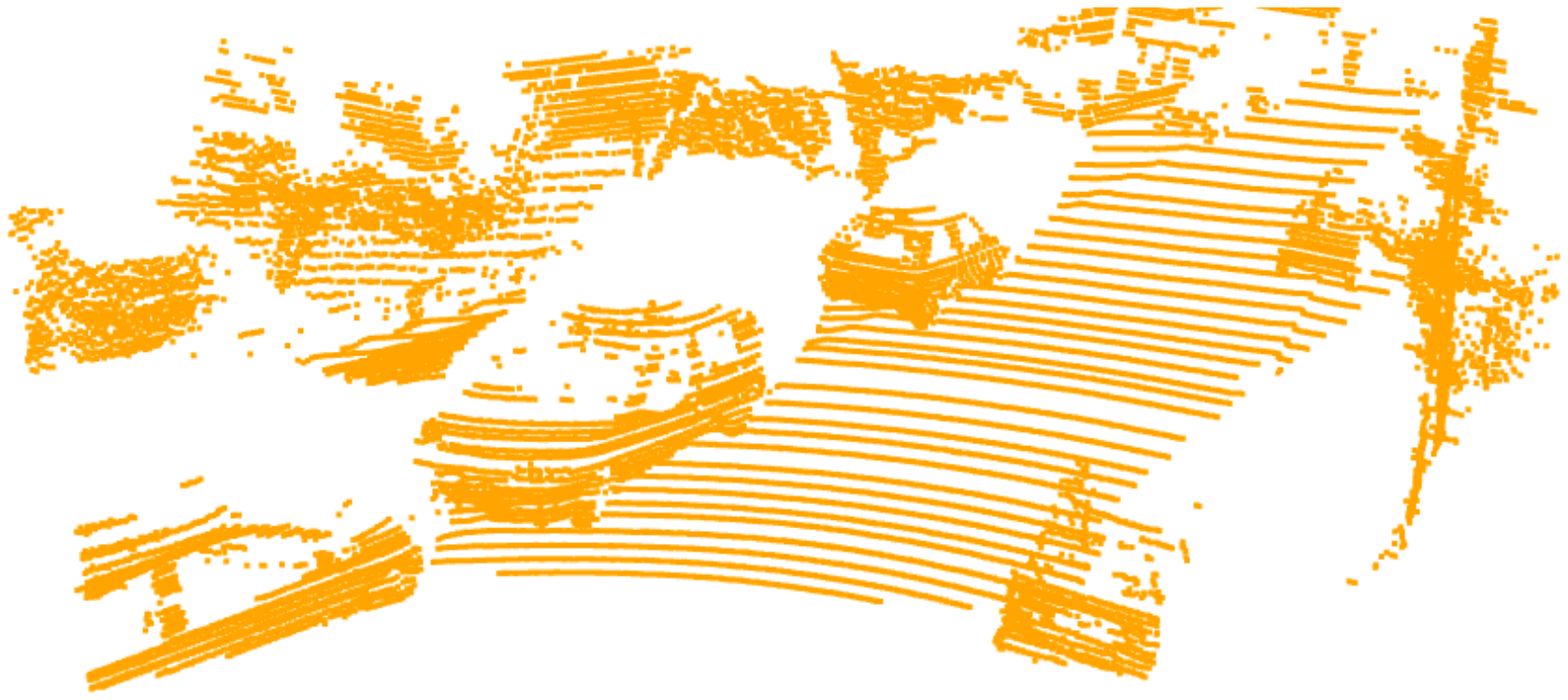}
        \caption{Original}
        \label{fig: case_a}
    \end{subfigure}
    \begin{subfigure}{0.18\linewidth}
        \centering
        \includegraphics[width=\textwidth]{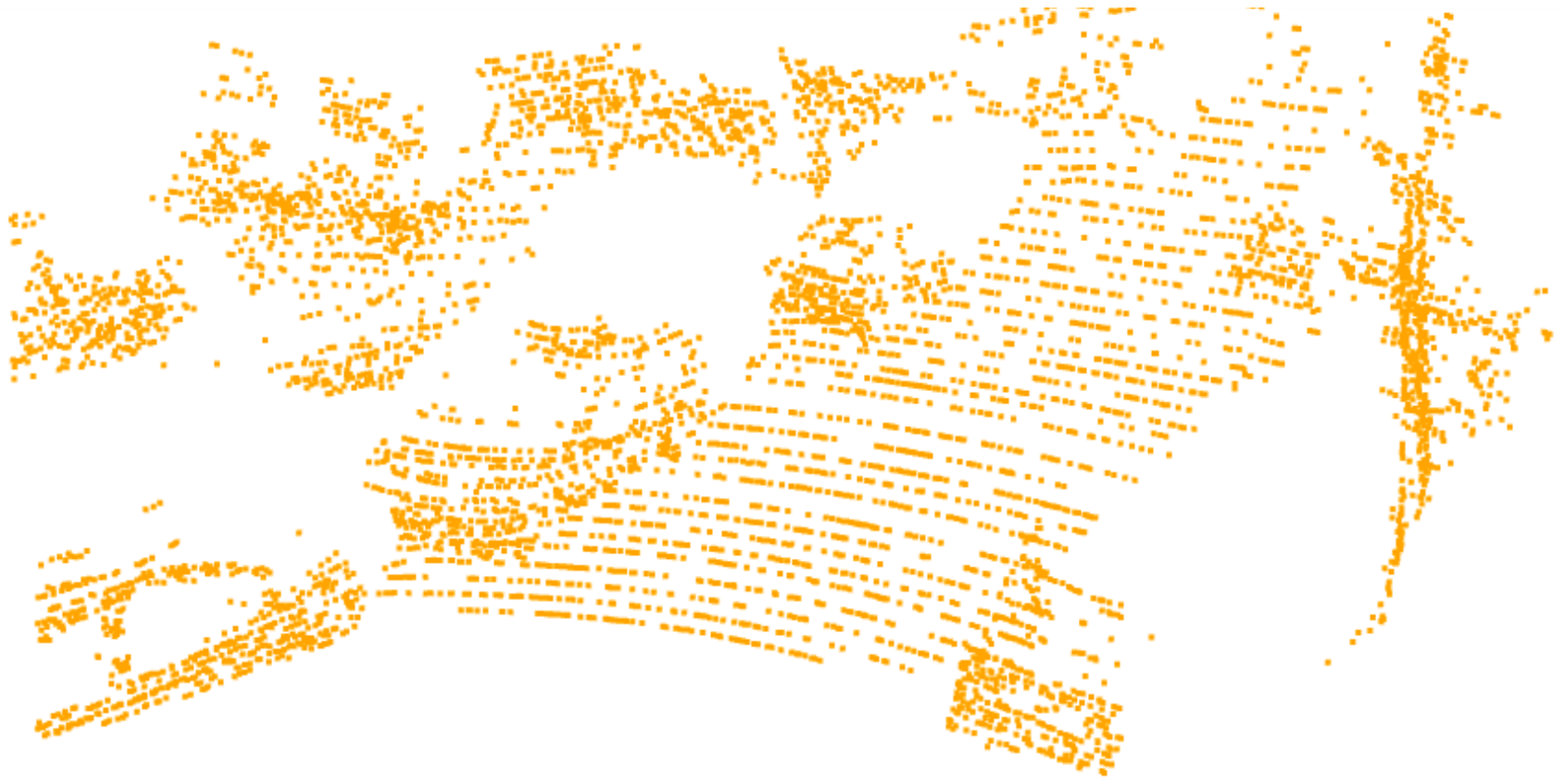}
        \caption{Random Sampling}
        \label{fig: case_b}
    \end{subfigure}
     \begin{subfigure}{0.18\linewidth}
        \centering
        \includegraphics[width=\textwidth]{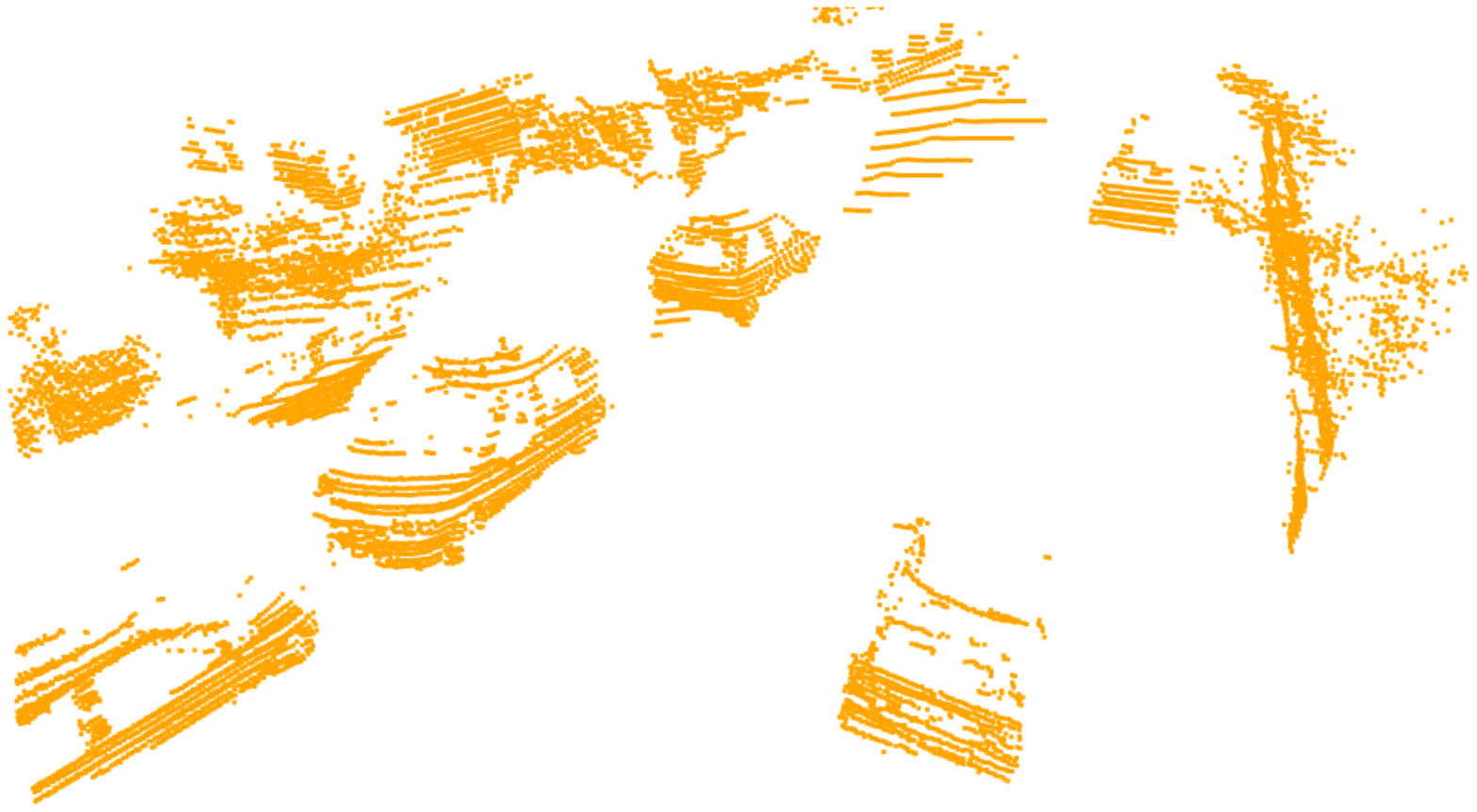}
        \caption{Object-aware Sampling}
        \label{fig: case_c}
    \end{subfigure}
    \caption{Demo of different sampling strategy with 50\% sample rate}
    \label{fig: case_study}
\end{figure*}

\subsection{Related Work}
\label{sec: related_work}

\noindent\textbf{Important Points Selection}.
To achieve the goal of selecting the representative points from a point cloud, deep learning-based methods~\cite{dovrat2019learning, lang2020samplenet} have been proposed. 
However, both methods have the limitation that they are designed for single-object point clouds and may struggle to extend to large-scale point clouds for outdoor scenes. 
Those methods use PointNet~\cite{qi2017pointnet} or PointNet++~\cite{qi2017pointnet++} as the backbone, 
which can hardly be extended to the outdoor point cloud feature extraction due to significant computational and memory requirements as mentioned in~\cite{zhou2018voxelnet}. It is limited by both the complexity and scale, and the training efficiency for the outdoor point clouds.
Another stream of studies integrates the point cloud subsampling as part of the deep learning module, which is trained end-to-end~\cite{zhang2022not, hu2020randla}.
Although such methods can handle large-scale outdoor point clouds, they have limited capability to achieve our desired feature of being compatible with different downstream models. The reason is that the conventional point selection methods inevitably have a bias towards selecting points based on the design of downstream model~\cite{glasmachers2017limits}.
For example, in~\cite{zhang2022not}, foreground points are deemed more important for the downstream model, leading
the point selection module to strategically pick those points, which has no guarantee of performance when the downstream model changes. In~\cite{wu2021redal}, a regional-based point selection using active learning is proposed. Although this study focuses on point cloud labelling, it conveys an essential idea that different objects in a scene may not require uniform point density for perception. In their experiments on point clouds for indoor scenes, it is obvious that removing the point of the floor has a very limited effect on the performance of the trained model. 

\noindent\textbf{Point Cloud Feature Extraction}. 
Learning from point clouds 
involves extracting richer information from the points, thereby facilitating downstream tasks.
Deep learning has been widely used in extracting features from point clouds, 
with methods broadly categorized
into point-based and voxel-based approaches. With pioneer works including \cite{qi2017pointnet, qi2017pointnet++, wang2019dynamic} utilize the convolutional network to points and break the ice for point-based solutions, the later studies provide more elegant solutions by incorporating more point features~\cite{
Wang2020feature, zhao2021point, wu2022point}. Compared to the point-based methods, the voxel-based methods are more efficient and computationally friendly~\cite{liu2019point, tang2020searching, zhu2021cylindrical}, which can be easily extended to the large-scale autonomous driving-related dataset \cite{kitti2019iccv, waymo2020CVPR, caesar2020nuscenes}. Recently, richer resources, such as camera images and radar data, have been introduced to facilitate feature learning. The most recent state-of-the-art studies are utilizing multiple resources to enhance point cloud perceptions ~\cite{yan2018second, yan20222dpass, zhang2023learning}.

\noindent\textbf{Compression}.~In the context of our goal to reduce data size, compression is a way to achieve it. From the literature, the compression can be divided into two categories: Octree-based methods \cite{schnabel2006octree, huang2020octsqueeze, que2021voxelcontext} and Range Image-based methods \cite{zhou2022riddle}. However, resorting to compressed data inevitably leads to the necessity for decompression. The uncompressed data maintains the original size and does not reduce the hardware demands for model training. Therefore, compression techniques do not align directly with our research focus, presenting a divergent path rather than a directly comparable methodology.

\section{Methodology}
\label{sec: method}

In this section, we first demonstrate a case study to explain the effectiveness of our Object Presence Detection based solution design (Section~\ref{sec: method_case_study}). Then, we formally introduce our solution. As shown in Figure~\ref{fig: overview}, our solution has two major steps. The first step is Object Presence Detection, which aims to develop a binary classifier to discern object points from background points. We propose two alternative solutions: the supervised density peak-based method (Section~\ref{sec:method_peak}) and the supervised Bayes enhanced method (Section~\ref{sec:method_bayes}).
The second step is Sample Budget Allocation,
with which we allocate sampling budgets to prioritize the object point set over the background point set (Section~\ref{sec:method_sampling}). 

\subsection{Case Study}
\label{sec: method_case_study}
In this case study, we use an example to illustrate the motivation of our solution design and its effectiveness. 
Intuitively, in a complete point cloud, the points of the objects (e.g., pedestrians, vehicles) hold more significance than the points of the background (e.g., ground and buildings). 

In our case study, we use two different sample strategies: (1) Random Sampling. (2) Object-aware Sampling, which is prone to retain more object points than the background points. We demonstrate the result of a 50\% sample rate.

\noindent\textbf{Visualization Result.} Figure~\ref{fig: case_study} visualizes the result of different sample strategies. According to the visualization, compared with the original point cloud (in Figure~\ref{fig: case_a}), it is evident that Object-aware Sampling (in Figure~\ref{fig: case_c}) provides a better depiction of the objects, whereas random sampling (in Figure~\ref{fig: case_b}) tends to blur the entire point cloud.

\noindent\textbf{Experimental Results.} We validate the effectiveness of the ``Object-aware Sampling'' strategy through simple experimental results. Table~\ref{demo: case_study} illustrated the inference results with the full point cloud and the two aforementioned sampling strategies (conducted with Part-A2~\cite{shi2020parta2} as the inference model). The results indicate that the mAP significantly drops when employing the random sampling strategy compared to the original full point cloud. However, with the ``Object-aware Sampling'', although there is a drop in mAP, it remains within an acceptable range, especially considering the halving of the number of points.



\begin{table}[h]
  \centering
  \small
  \caption{Eval mAP of various sample strategies on KITTI}
  \begin{tabular}{
    l
    S[table-format=2.2]
    S[table-format=2.2]
    S[table-format=2.2]
  }
    \toprule
    {Strategy} & {Easy} & {Med} & {Hard} \\
    \midrule
    Original & 77.84 & 64.45 & 60.71 \\
    Random   & 70.79 & 56.57 & 53.09 \\
    Keep Object & 76.07 & 63.00 & 59.27 \\
    \bottomrule
  \end{tabular}
  \label{demo: case_study}
\end{table}

\subsection{Object Presence Detection with Density Peak}
\label{sec:method_peak}
In this section, we introduce an efficient statistics-based object presence detection approach, to achieve our goal of classifying points into two categories: object points and background points. 
Essentially,
this method provides the swift identification of object presence while eliminating the need for training.

\begin{figure}[h]
    \centering
    \includegraphics[width=0.48\textwidth]{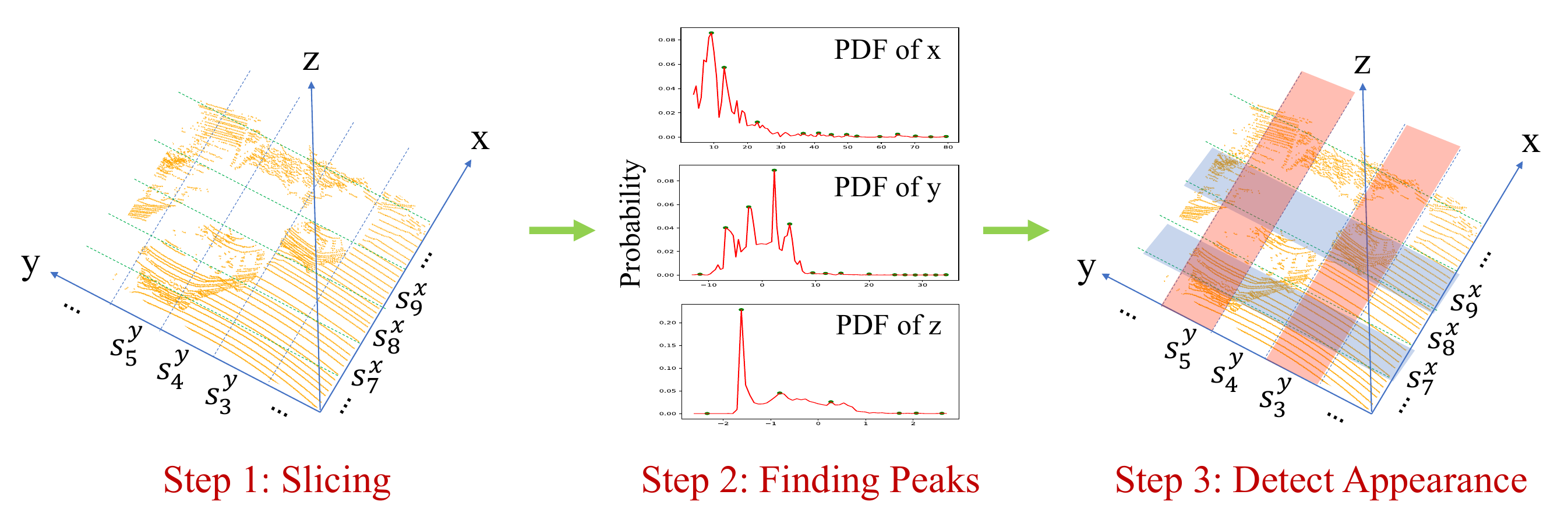}
    \caption{Demonstration of Density Peaks based classification}
    \label{fig: method}
\end{figure}


With the 3D features of the point cloud, we can obtain information from statistics of $x$, $y$, and $z$ axes. We utilize the density peak as the statistical feature of the three axes to detect the presence of the object, as demonstrated in Figure \ref{fig: method}. To elaborate, our method mainly comprises three steps:

\noindent\textbf{Step 1: Slicing.} The purpose of slicing is to get initial statistics from the point cloud for $x$, $y$, and $z$ axes respectively. Given a point cloud $\pointCloud$  that contains a set of points $\{p_1, p_2,..,p_{|P|}\}$ $\in \pointCloud$. Each point has three coordinates, i.e., ($p_i.x, p_i.y, p_i.z$), to represent its location in the point cloud. We evenly slice $\pointCloud$ along the X-axis and Y-axis into $m$ and $n$ slices respectively. We use $s^x_i$ to denote the $i_{th}$ slice in the $x$ direction. Thus, after slicing, we will have ${s^x_1, s^x_2, ..., s^x_m}$ by slicing  the X-axis, and we will get ${s^y_1, s^y_2, ..., s^y_n}$ by slicing the Y-axis (as shown in Figure \ref{fig: method} Step 1). 

It is worth noting that we do not perform the slicing along the Z-axis because it lacks sufficient features that contribute to Step 2, and it may negatively impact the completeness of the object. For details on how to determine the number of slices expected for each axis, please refer to ``Data-driven Parameter Study" in our technical report~\cite{mygit}.

\noindent\textbf{Step 2: Finding Local Peaks.} Since the point cloud uses points to formulate objects, the appearance of objects inevitably results in a surge in the number of points along both the X-axis and Y-axis. By leveraging this property, we can identify the slices that are more likely to contain objects.

Within each slice $s$, we can calculate the number of points it contains, thereby determining the point density of the corresponding slice. 
This point density is a crucial feature we use to detect the presence of objects. 
Considering the points tend to be sparse in distance places, instead of identifying the overall peak, we focus on detecting the local peaks of point density (as shown in Figure \ref{fig: method} Step 2). 
By specifying a stride, we iterate through the data and find the peak within the stride as the local peak. The detailed algorithm can be found in ``Local Peak Search Algorithm" of our technical report~\cite{mygit}.

\noindent\textbf{Step 3: Detecting Object Presence.} From the previous step, we identify the slices that possibly contain objects along both the X-axis and Y-axis. To pinpoint the presence of the object, we examine the intersections of the slices along the X-axis and Y-axis. We designate the points within these intersections as significant, while considering the remaining points as less significant background points. 

We use an example to illustrate object presence detection, as shown in Figure \ref{fig: method}.
In Step 2, peaks were identified in slices $s^x_7, s^x_9$ along the X-axis and $s^y_3, s^y_5$ along the Y-axis. Subsequently, in step 3, we categorize the points falling within the intersections of ($s^x_7, s^y_3$), ($s^x_7, s^y_5$), ($s^x_9, s^y_3$) and ($s^x_9, s^y_5$) as object points, while the remaining points are considered background points. 

\noindent\textbf{Special Filter: Z-axis.} The Z-axis filter plays a unique role within the filtering process. While the Z-axis is relatively less sensitive to object appearances, it has a distinctive ability to effectively identify a critical subset of background points—the ground. Despite the points for the ground has limited contribution to object detection, it often constitutes a substantial portion of a point cloud. By analyzing the Z-axis values, we can identify a global peak, which can be confidently classified as background points.

\subsection{Object Presence Detection with Na\"ive Bayes}
\label{sec:method_bayes}
In Section~\ref{sec:method_peak}, we introduce the unsupervised density peak-based detection approach. This method is superior in efficiency but at the cost of effectiveness. 
Therefore, in this section, we propose an effective supervised solution by leveraging the Na\"ive Bayes to improve the learning of point density while improving performance.
Using the training data, we build the Bayes model for each cross-region to predict the probability of
objects appearing in that specific region.
\subsubsection{Na\"ive Bayes Based Binary Classification}

The Na\"ive Bayes-based binary classification is extended from the previous density peak-based method in Section \ref{sec:method_peak} by incorporating training data to develop a Na\"ive Bayes model. It enables a more precise assessment of object presence. The Na\"ive Bayes-based method comprises two fundamental steps as follows: 

\noindent\textbf{Step 1. Slicing and Crossing.} We use the same slicing strategy as introduced in Section \ref{sec:method_peak}, which slices a point cloud \pointCloud into \binX \ slices along X-axis and \binY \ slices along Y-axis. However, unlike dynamically estimating \binX \ and \binY \ for each point cloud in the peak-based method, with the Na\"ive Bayes-based method, we fix \binX \ and \binY \ for all point clouds $P \in \mathcal{P}$. By crossing slices, we get $m \times n$ regions.

\noindent\textbf{Step 2. Building Na\"ive Bayes Models for classification.}
With the aforementioned steps, the point cloud \pointCloud \ is divided into $m \times n$ regions. Each region results from the intersection of slice $s^x_i$ and slice $s^y_j$, which is denoted as $r_{(s^x_i, s^y_j)}$. For each region, we use Na\"ive Bayes to train a model over the training set $\mathcal{P}_{train}$. For example, in region $r_{(s^x_i, s^y_j)}$, given its point density in $s^x_i$, $s^y_j$ (which is denoted by $r.d_{x}$ and $r.d_{y}$), we use equation~\ref{eq: bayes} to train the corresponding Bayesian model. In total, we would have $m \times n$ Bayesian models. It is worth noting that, this is a binary classification task, intending to categorise points into object point sets or background point sets. To prepare the training data and the detailed explanation of the equation, please refer to `` Prepare Training Data for Na\"ive Bayes" and ``Na\"ive Bayes Explanation" of our technical report~\cite{mygit}.
\begin{equation}
    P(y_{r}^{obj}|r.d_{x}, r.d_{y}) = \frac{P(y_{r}^{obj})\times P(r.d_{x}, r.d_{y}|y_{r}^{obj})}{P(r.d_{x}, r.d_{y})} 
\label{eq: bayes}
\end{equation}

\subsubsection{Na\"ive Bayes Models as Filter}

To achieve the goal of differentiating object points from background points, we utilize our trained $m\times n$ Na\"ive Bayes model to filter out background points with higher precision. Compared with the density peak-based method which is very efficient, the Na\"ive Bayes enhanced detector is slower because using the model for prediction is time-consuming. To ensure both effectiveness and efficiency, we create a staged filter:

\noindent\textbf{a) Coarse-grained Filtering Stage.} As objects in the point cloud for a scene are sparsely distributed, there is a high probability that many regions would never have objects present. Such regions lack positive examples for the density of objects appearing. Therefore, we can confidently predict a high likelihood of these regions being environmental regions, even without invoking the model for predictions.

\noindent\textbf{b) Fine-grained Filtering Stage.} Built upon the efficiency gains achieved through the coarse-grained filter, this step focuses on the remaining regions. We leverage the previously mentioned Na\"ive Bayes models to predict the categorization outcomes of the remaining regions.

\subsection{Sampling}
\label{sec:method_sampling}
To ensure both efficiency and effectiveness, we perform the sampling with strategies including imbalanced sample budget allocation and samples with different granularity. Essentially, the object points hold more significance compared to the background points. Thus, we retain more object points and sampled with finer granularity, while allocating less budget to background points and sampled with coarse granularity to accelerate the efficiency. The detailed implementation is as follows:

\noindent\textbf{a) Sampling Budget Allocation.} Given an overall sampling rate, we can calculate the anticipated amount of points to be sampled. With this budget, our strategy involves a deliberate allocation of more budget to object points, while reserving a smaller portion to select background points. 

\noindent\textbf{b) Sampling with Different Granularity.} Given the relatively lower significance of the environmental point set, we use the efficient random sampling strategy to select points from this set. For the object point set, we will perform region-based random sampling. Specifically, we execute random sampling within each distinct region containing objects instead of an overall random sampling over the whole set. Although this may sacrifice some efficiency, the object points are preserved to the utmost extent possible.

\section{Experiments}

\begin{figure*}[!t]
    \centering
    \begin{subfigure}{0.225\linewidth}
        \centering
        \includegraphics[width=\textwidth]{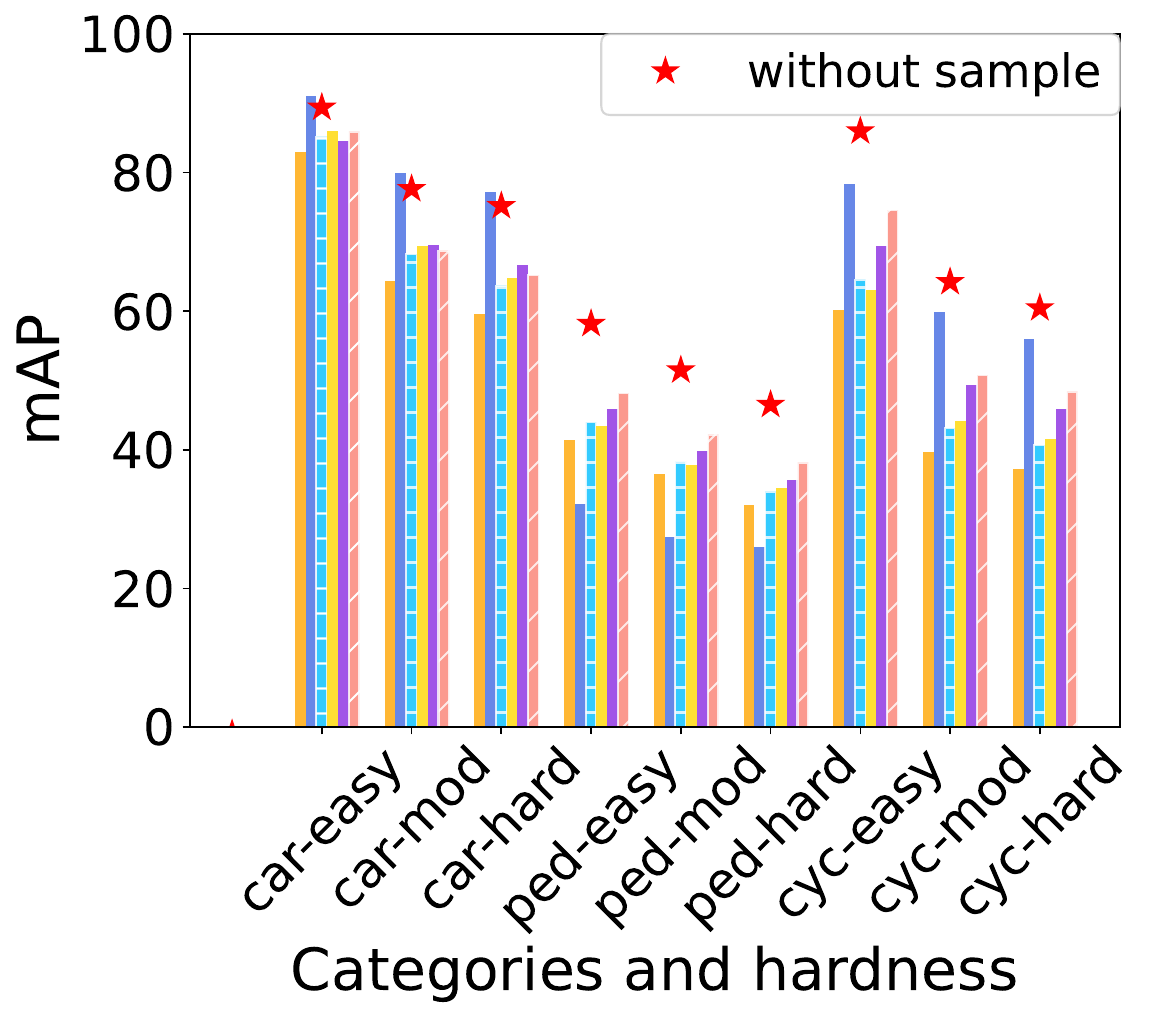}
        \caption{Sample Rate 30\%}
        \label{fig1: a}
    \end{subfigure}
    \begin{subfigure}{0.21\linewidth}
        \centering
        \includegraphics[width=\textwidth]{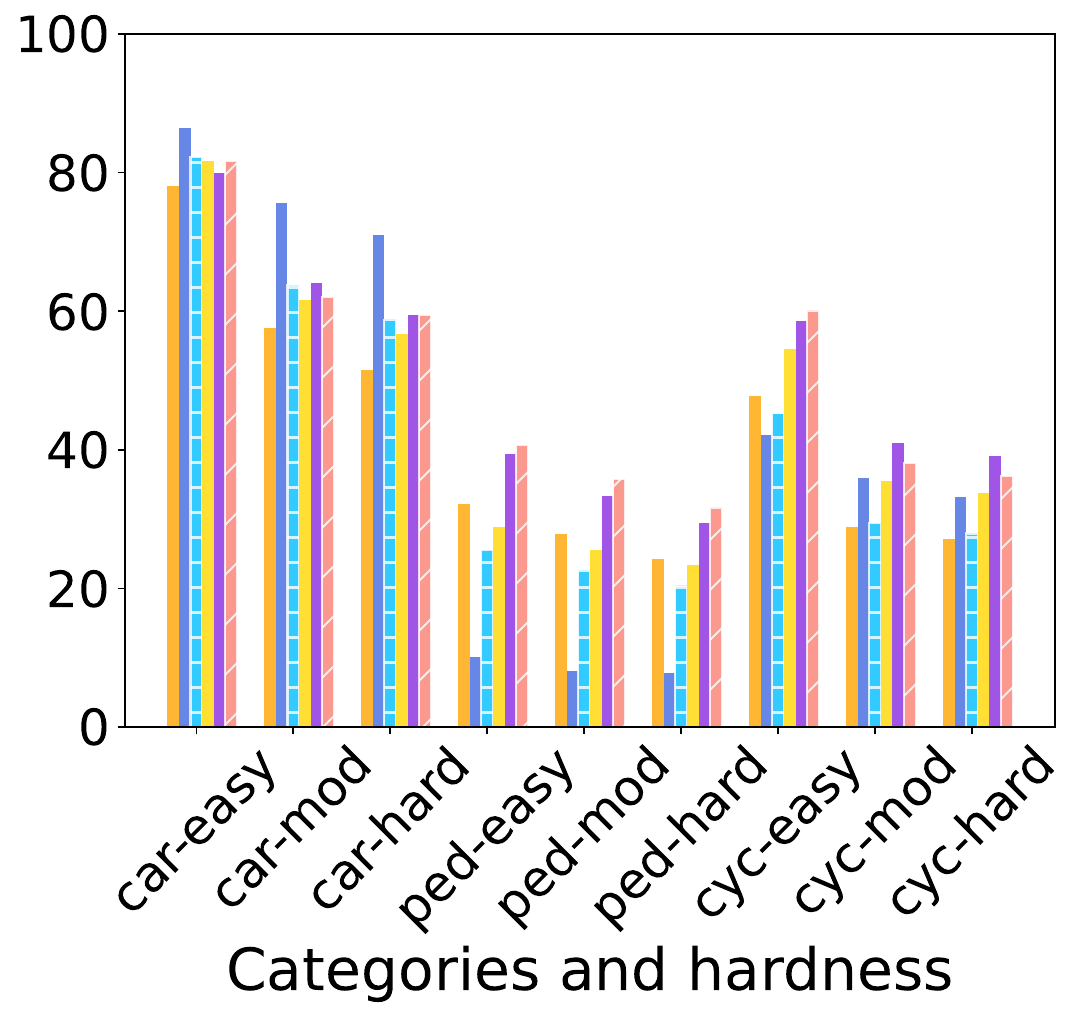}
        \caption{Sample Rate 20\%}
        \label{fig1: b}
    \end{subfigure}
    \begin{subfigure}{0.21\linewidth}
        \centering
        \includegraphics[width=\textwidth]{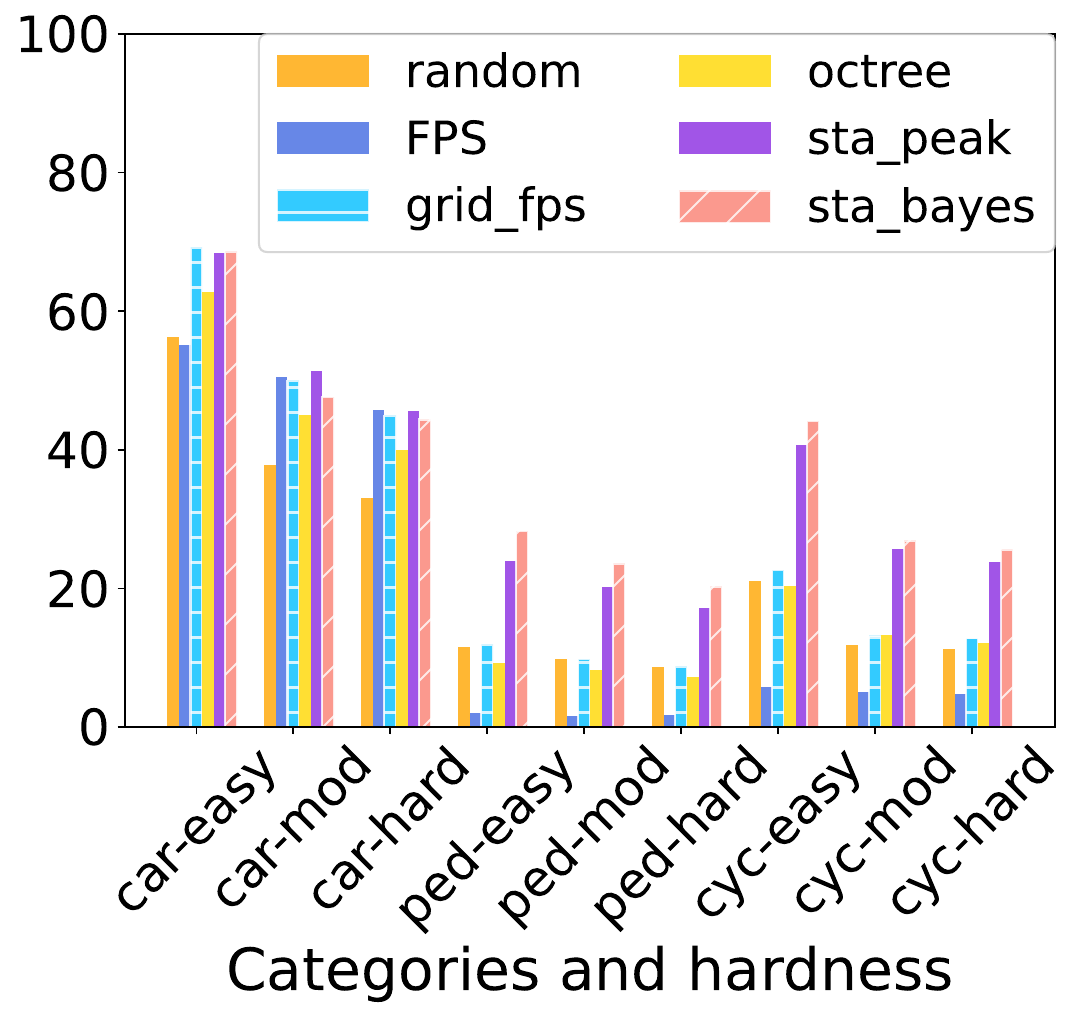}
        \caption{Sample Rate 10\%}
        \label{fig1: c}
    \end{subfigure}
    \caption{Categorical mAP with Part-A2 on KITTI}
    \label{fig1}
\end{figure*}

\begin{figure*}[!t]
    \centering
    \begin{subfigure}{0.225\linewidth}
        \centering
        \includegraphics[width=\textwidth]{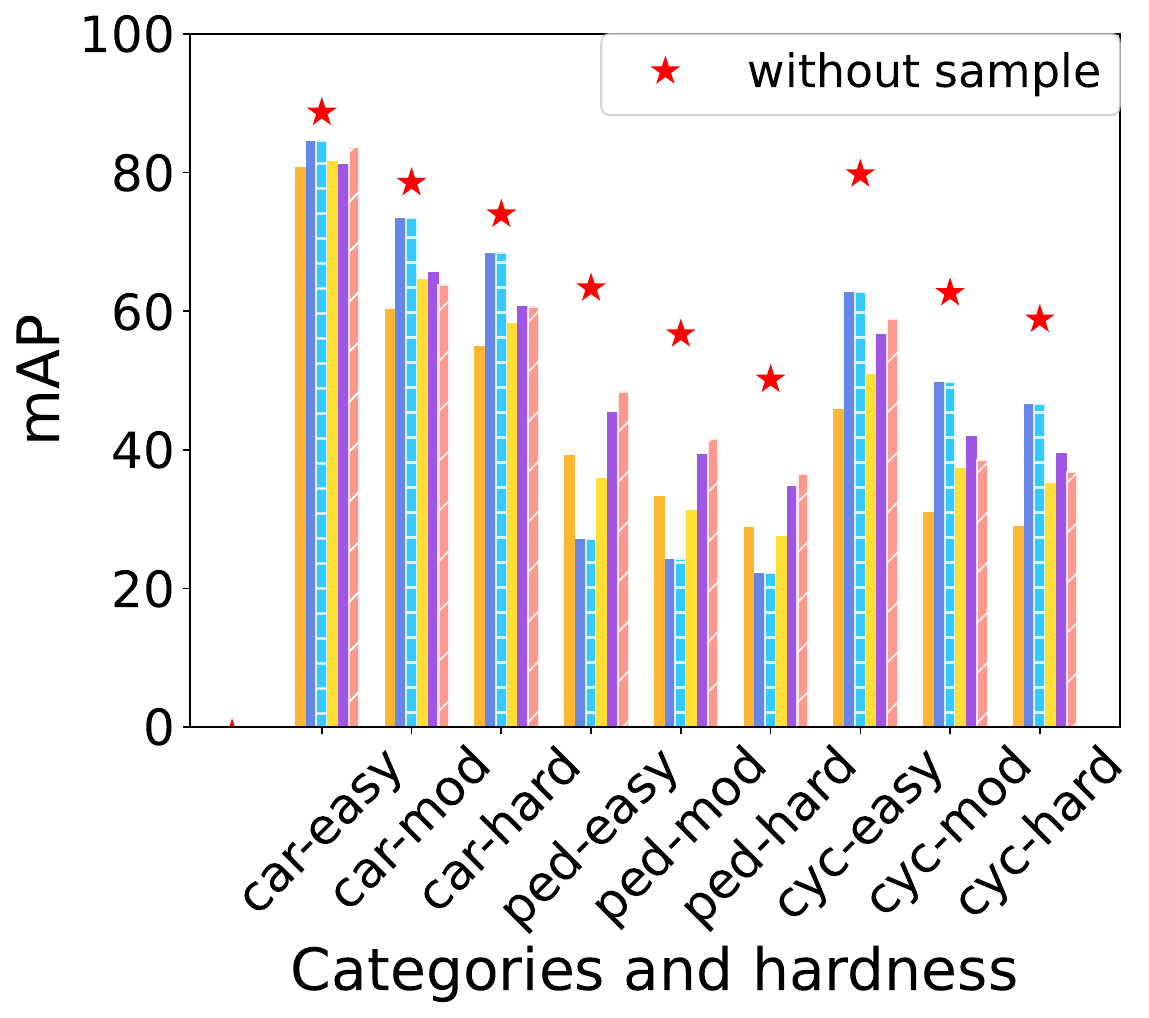}
        \caption{Sample Rate 30\%}
        \label{fig3: a}
    \end{subfigure}
    \begin{subfigure}{0.21\linewidth}
        \centering
        \includegraphics[width=\textwidth]{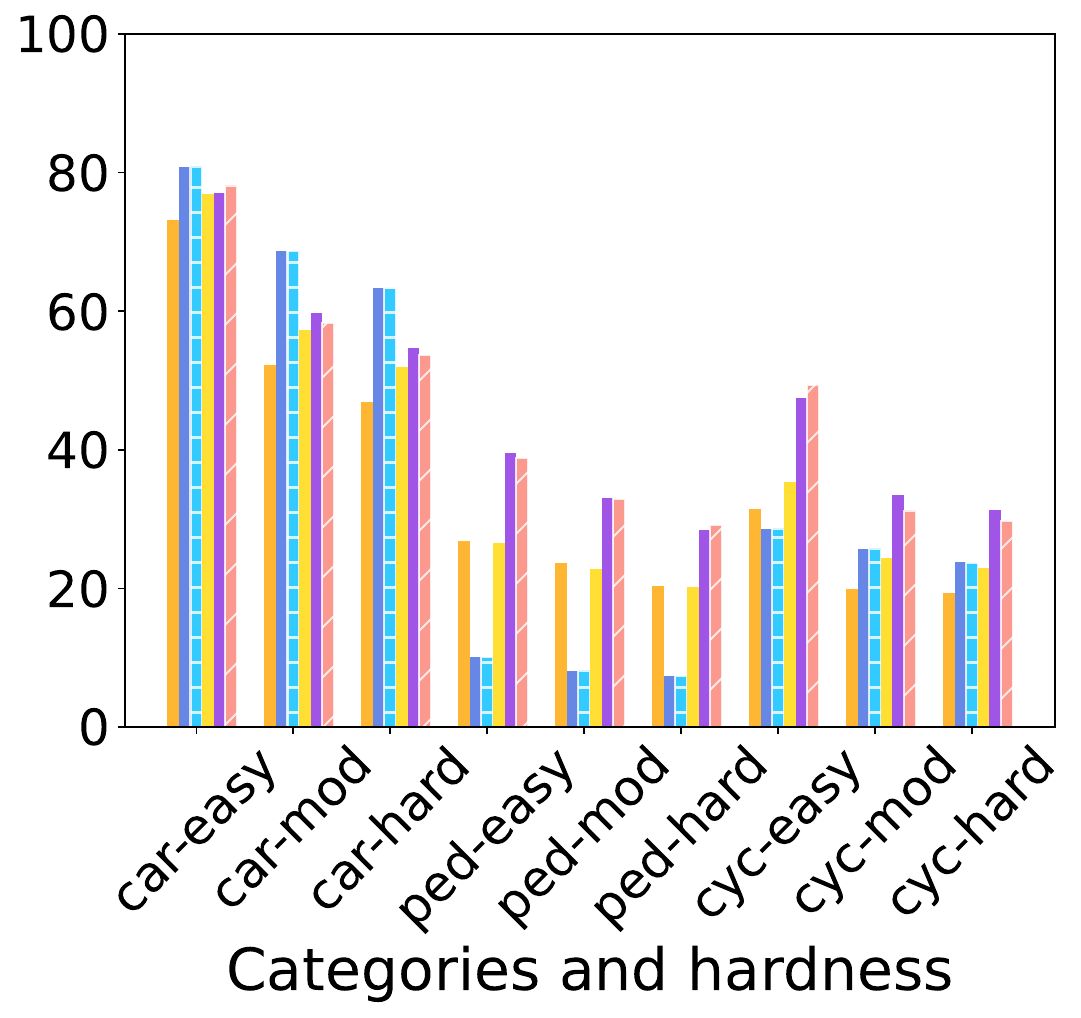}
        \caption{Sample Rate 20\%}
        \label{fig3: b}
    \end{subfigure}
    \begin{subfigure}{0.21\linewidth}
        \centering
        \includegraphics[width=\textwidth]{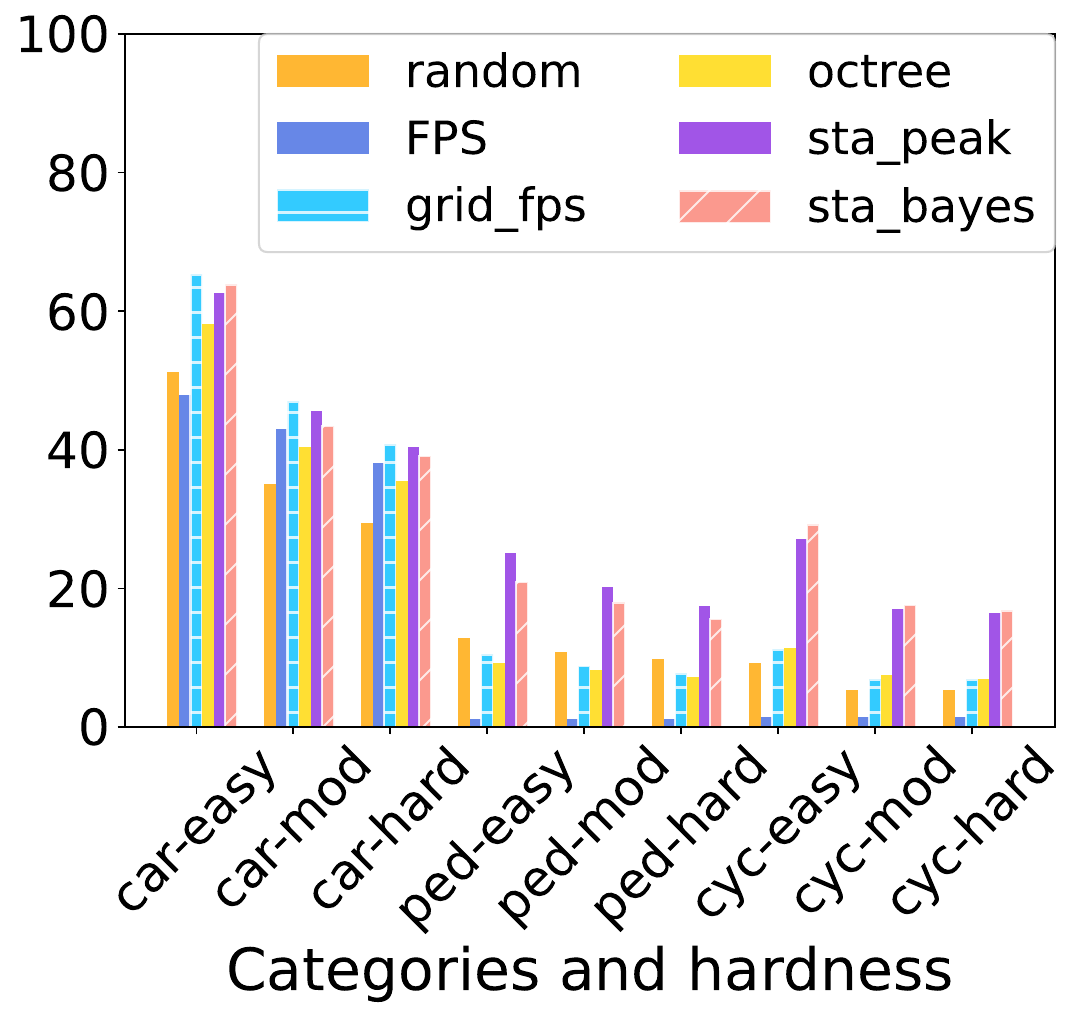}
        \caption{Sample Rate 10\%}
        \label{fig3: c}
    \end{subfigure}
    \caption{Categorical mAP with SECOND on KITTI}
    \label{fig3}
\end{figure*}
\begin{figure*}[!t]
    \centering
    \begin{subfigure}{0.225\linewidth}
        \centering
        \includegraphics[width=\textwidth]{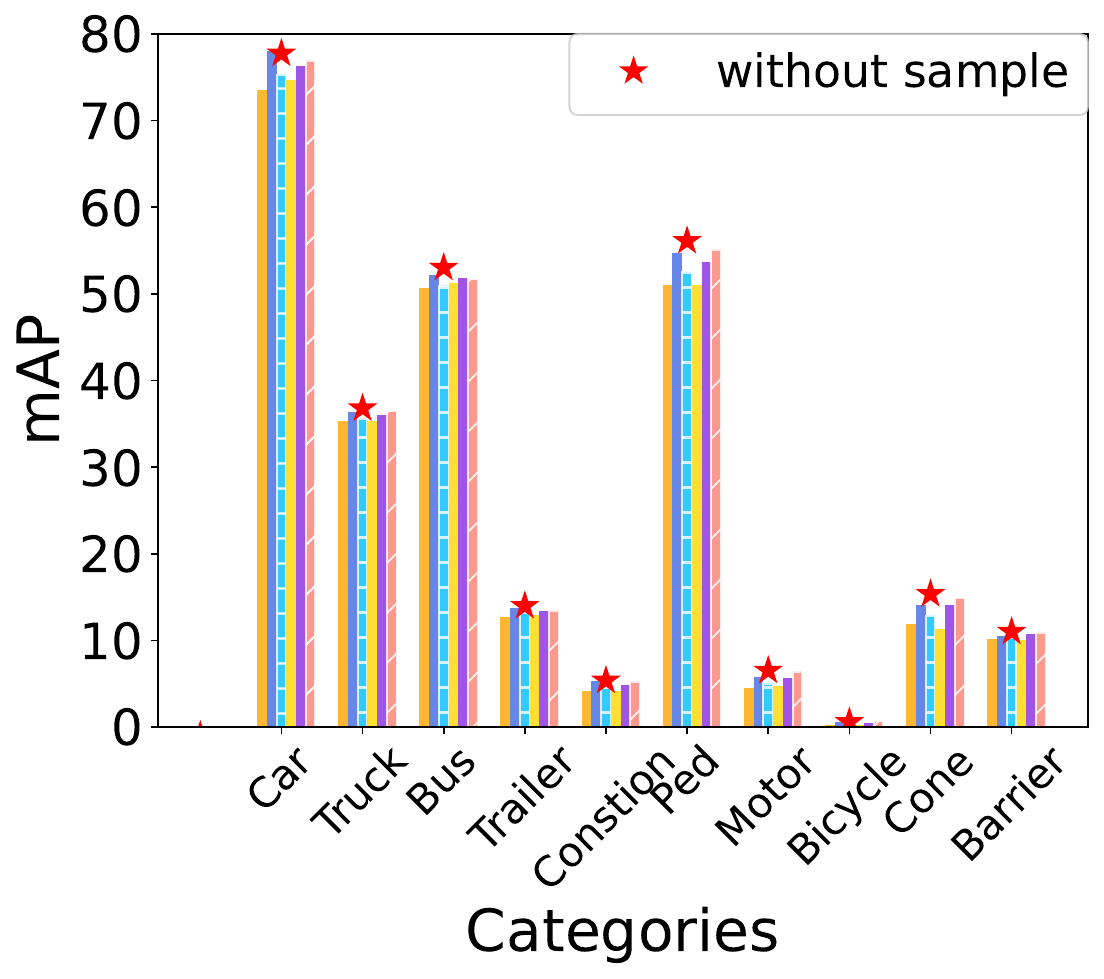}
        \caption{Sample Rate 10\%}
        \label{fig4: a}
    \end{subfigure}
    \begin{subfigure}{0.21\linewidth}
        \centering
        \includegraphics[width=\textwidth]{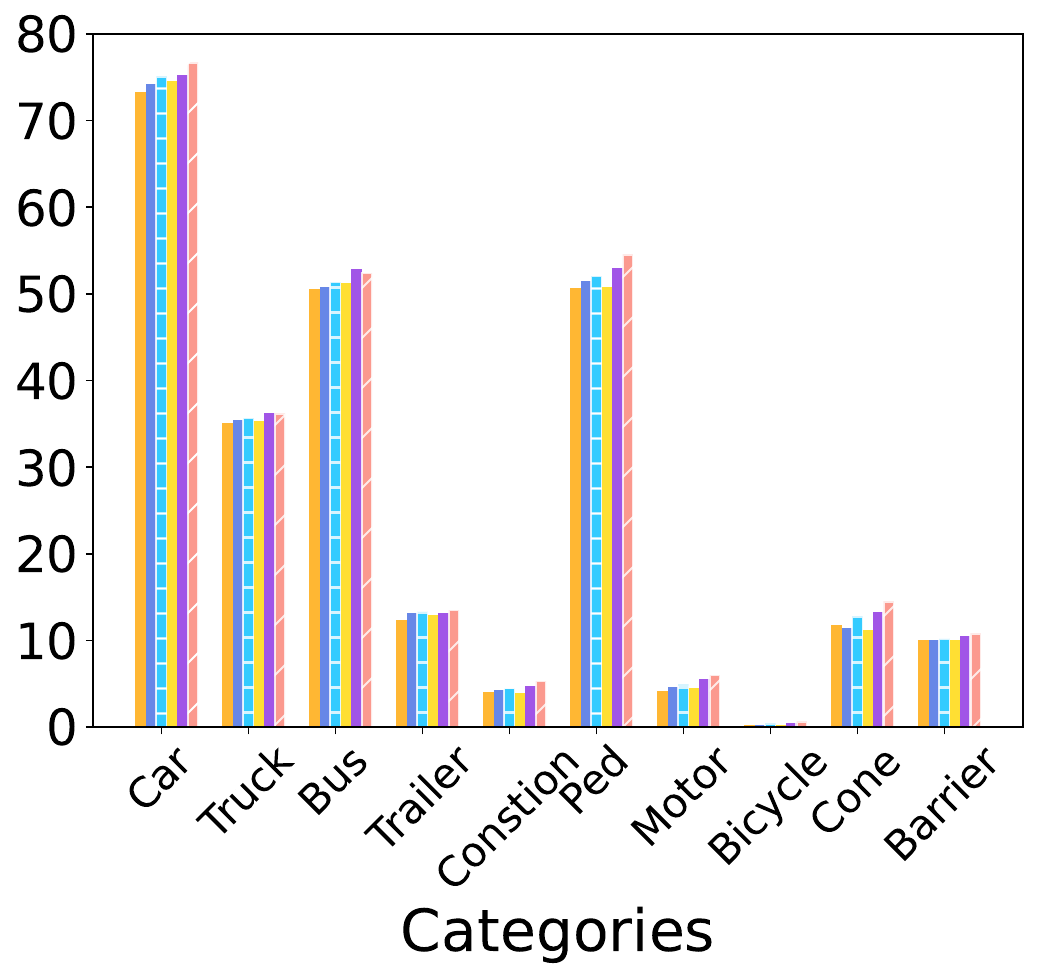}
        \caption{Sample Rate 8\%}
        \label{fig4: b}
    \end{subfigure}
    \begin{subfigure}{0.21\linewidth}
        \centering
        \includegraphics[width=\textwidth]{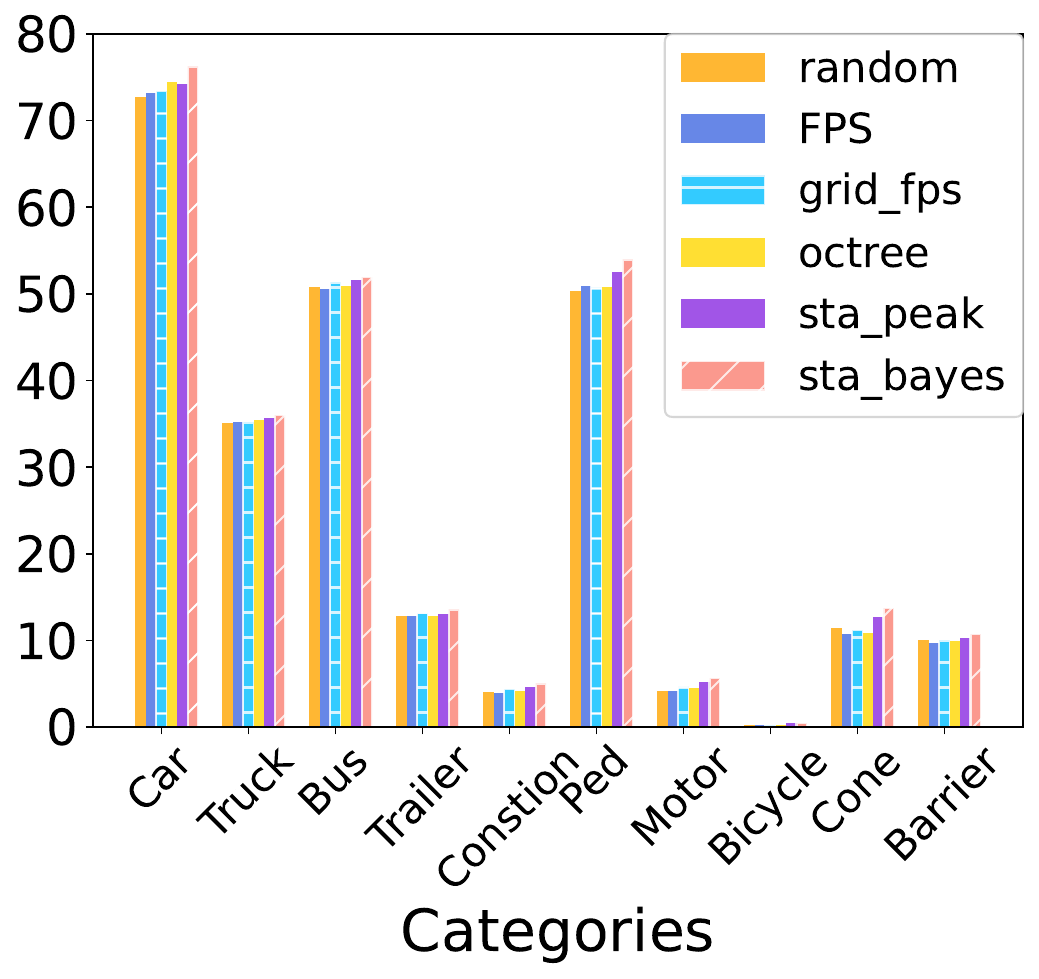}
        \caption{Sample Rate 6\%}
        \label{fig4: c}
    \end{subfigure}
    \caption{Categorical mAP with PointPillars on nuScenes}
    \label{fig4}
\end{figure*}
To evaluate the performance of the proposed approach, we conducted experiments on both KITTI and nuScenes datasets with different inference models. 
The effectiveness and efficiency of our sampling strategy are evaluated, with an additional ablation study conducted.

\subsection{Experiment Setup} 
\label{exp: setup}
All experiments were conducted on a server running Ubuntu 22.04.1, with Intel(R) Xeon(R) CPU E5-2697A v4 @ 2.60GHz and RTX2070. The code was implemented with the MMDetection3D Framework \cite{mmdet3d2020}, and can be found in the git repository~\cite{mygit}. We conducted experiments to investigate the following research questions:

\noindent\textbf{Q1.} How is the effectiveness of our proposed two methods compared with baselines? (Section \ref{exp: result_effectiveness}). 

\noindent\textbf{Q2.} What advantage does sta\_bayes have compared to sta\_peak? (Section \ref{exp: peak_vs_bayes} (a)) What is the generalization ability of our methods? (Section \ref{exp: peak_vs_bayes} (b)). 

\noindent\textbf{Q3.} What efficiency advantage does our proposed method have compared with the baselines? (Section \ref{exp: result_efficiency})  

\noindent\textbf{Q4.} What benefits do we obtain from different sampling budget allocations? How much does the Z-axis filter contribute to the sampling results? (Section \ref{sec: exp_ablation})

\noindent\textbf{Q5.} How do our methods compare to the deep learning approach? (See "Comparison with Trained MLP (Q5)" in the technical report~\cite{mygit})

\subsubsection{Datasets and Inference Models}

The experiment was conducted with KITTI dataset~\cite{kitti2019iccv}, and we selected Part-A2~\cite{shi2020parta2} and SECOND~\cite{yan2018second} as our target model for the downstream object detection task. We also conducted experiments on nuScenes~\cite{caesar2020nuscenes} and used PointPillars~\cite{lang2019pointpillars} as the inference model.

\subsubsection{Methods for Comparison}


\noindent\textbf{1) Random sampling.} This method involves randomly selecting points from the provided point cloud.

\noindent\textbf{2) Farthest Point Sampling (FPS).} FPS is implemented by selecting a subset of points that are maximally distant from one another.

\noindent\textbf{3) FPS with Regular grid sampling (grid\_fps).} We developed this baseline to address FPS limitations. It involves dividing the point cloud into grids and then applying FPS to select points from each grid cell.

\noindent\textbf{4) Octree sampling.} Octree sampling employs an Octree structure to partition space. The sampling technique is based on the approach in~\cite{el2018plane}.

\noindent\textbf{5) Statistics with Density Peak (sta\_peak).} The proposed method described in Section~\ref{sec:method_peak}.

\noindent\textbf{6) Statistics with Naive Bayes (sta\_bayes).} The proposed method described in Section~\ref{sec:method_bayes}. 
\begin{figure*}[!t]
    \centering
    \begin{subfigure}{0.19\linewidth}
        \centering
        \includegraphics[width=\textwidth]{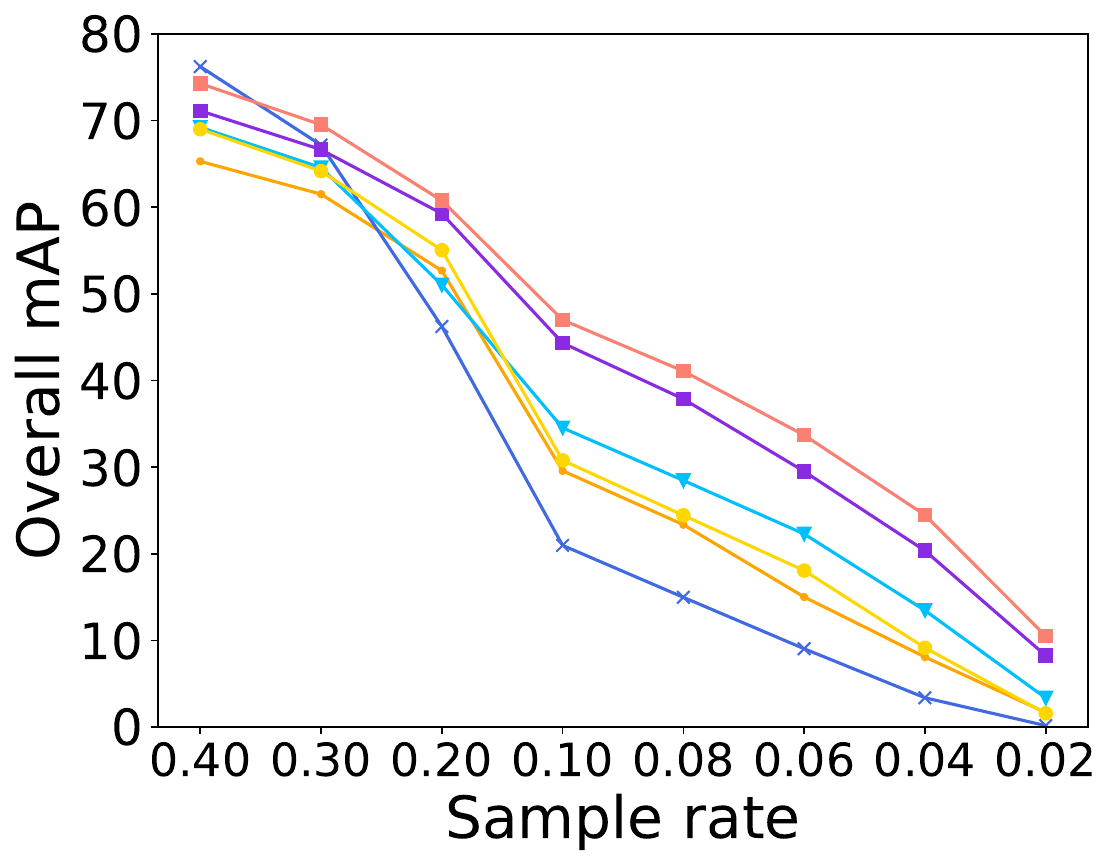}
        \caption{Easy}
        \label{fig5: a}
    \end{subfigure}
    \begin{subfigure}{0.18\linewidth}
        \centering
        \includegraphics[width=\textwidth]{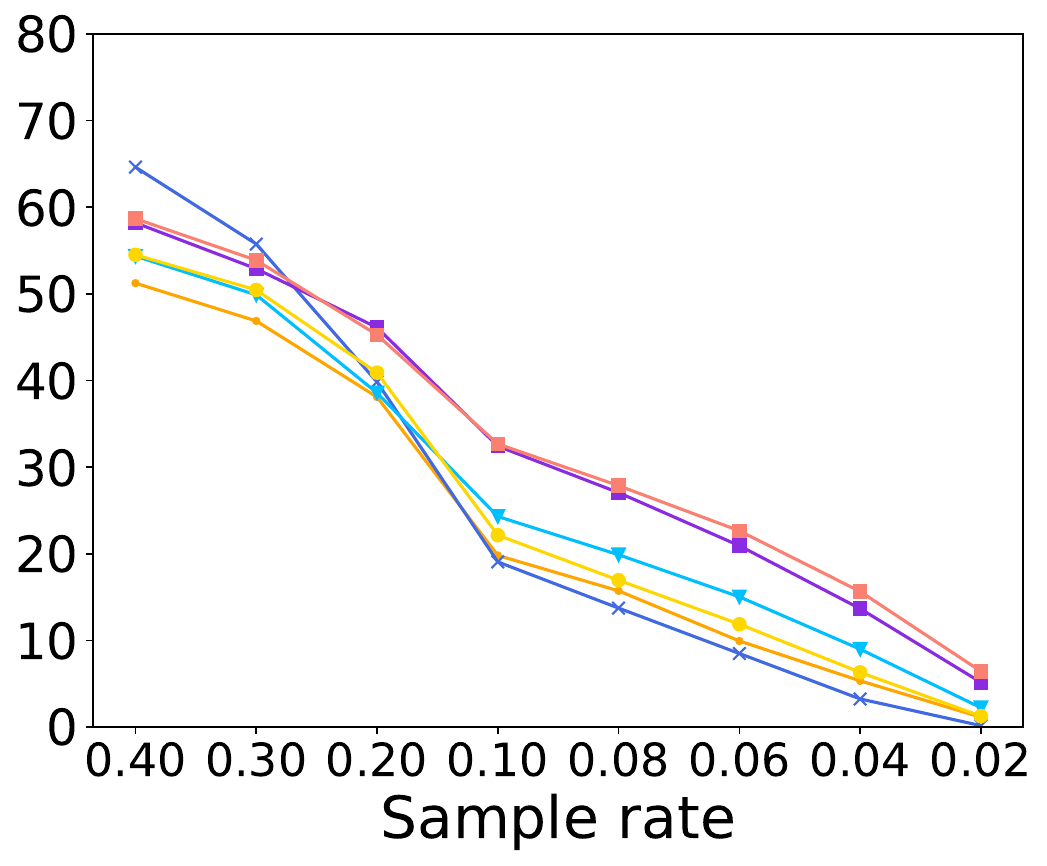}
        \caption{Mod}
        \label{fig5: b}
    \end{subfigure}
    \begin{subfigure}{0.18\linewidth}
        \centering
        \includegraphics[width=\textwidth]{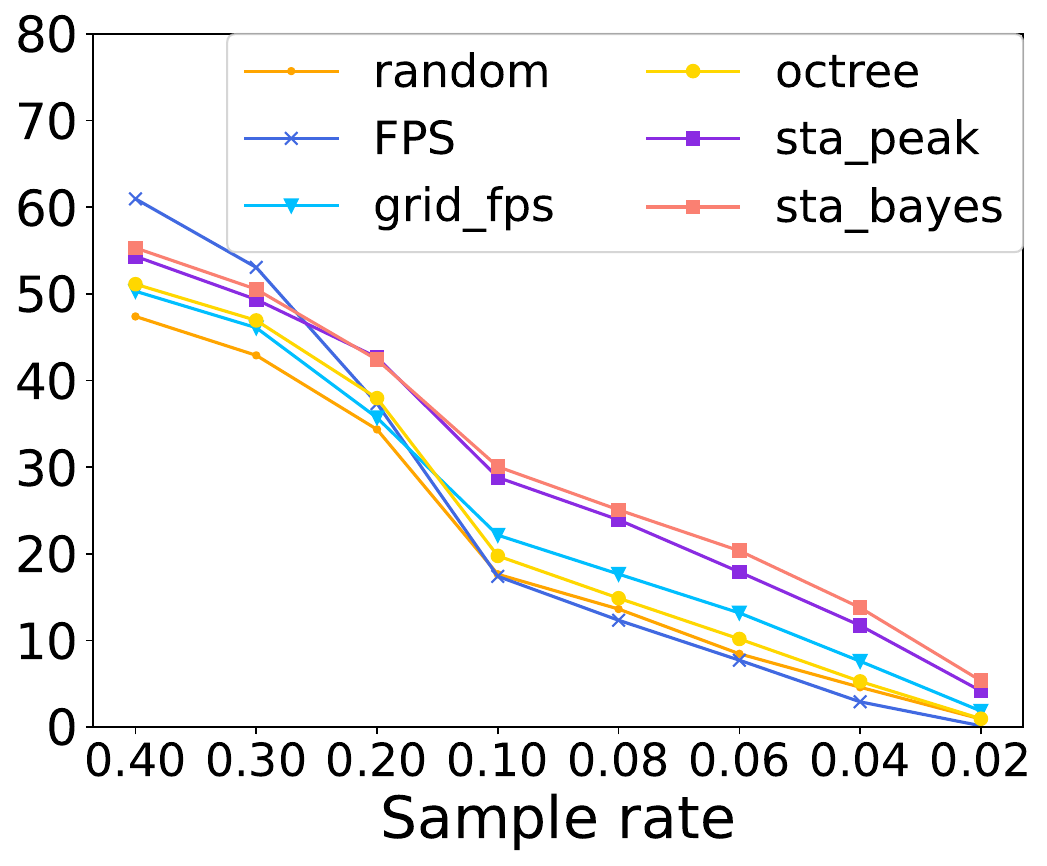}
        \caption{Hard}
        \label{fig5: c}
    \end{subfigure}
    \caption{Macro mAP with KITTI (Part-A2) }
    \label{fig5}
\end{figure*}
\begin{figure*}[!t]
    \centering
    \begin{subfigure}{0.19\linewidth}
        \centering
        \includegraphics[width=\textwidth]{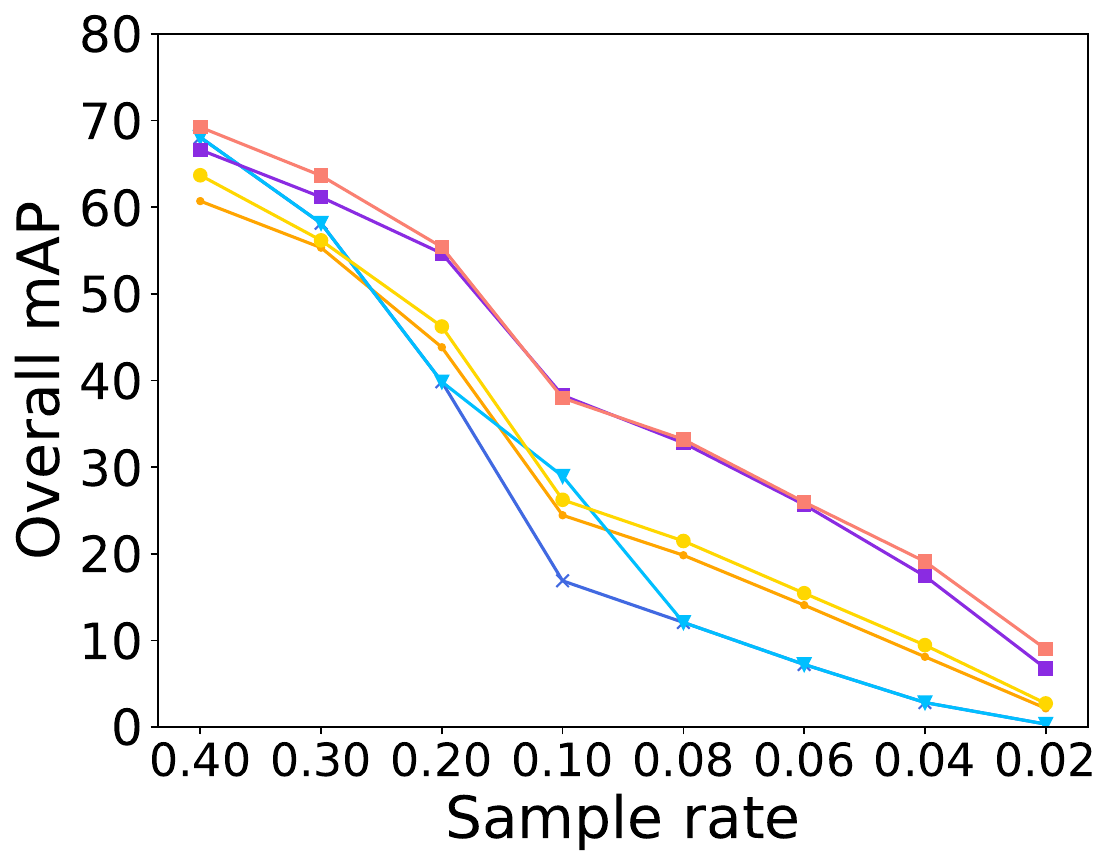}
        \caption{Easy (SECOND)}
        \label{fig5: d}
    \end{subfigure}
    \begin{subfigure}{0.18\linewidth}
        \centering
        \includegraphics[width=\textwidth]{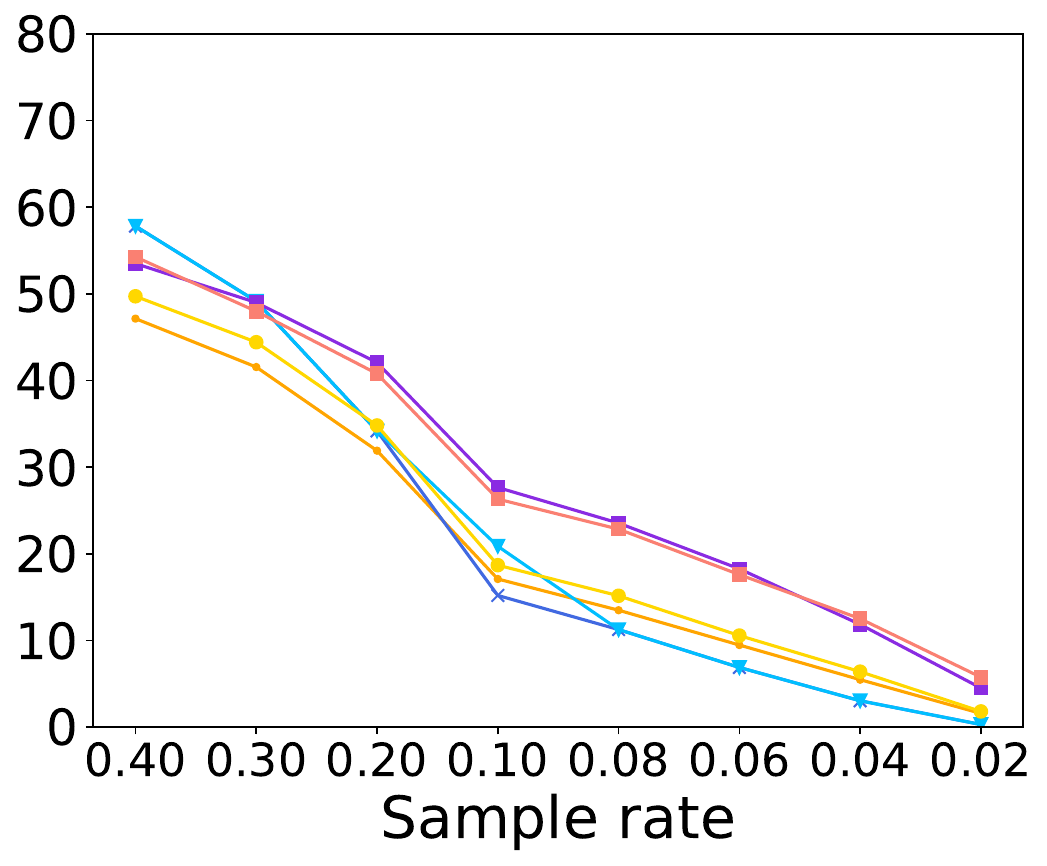}
        \caption{Mod (SECOND)}
        \label{fig5: e}
    \end{subfigure}
    \begin{subfigure}{0.18\linewidth}
        \centering
        \includegraphics[width=\textwidth]{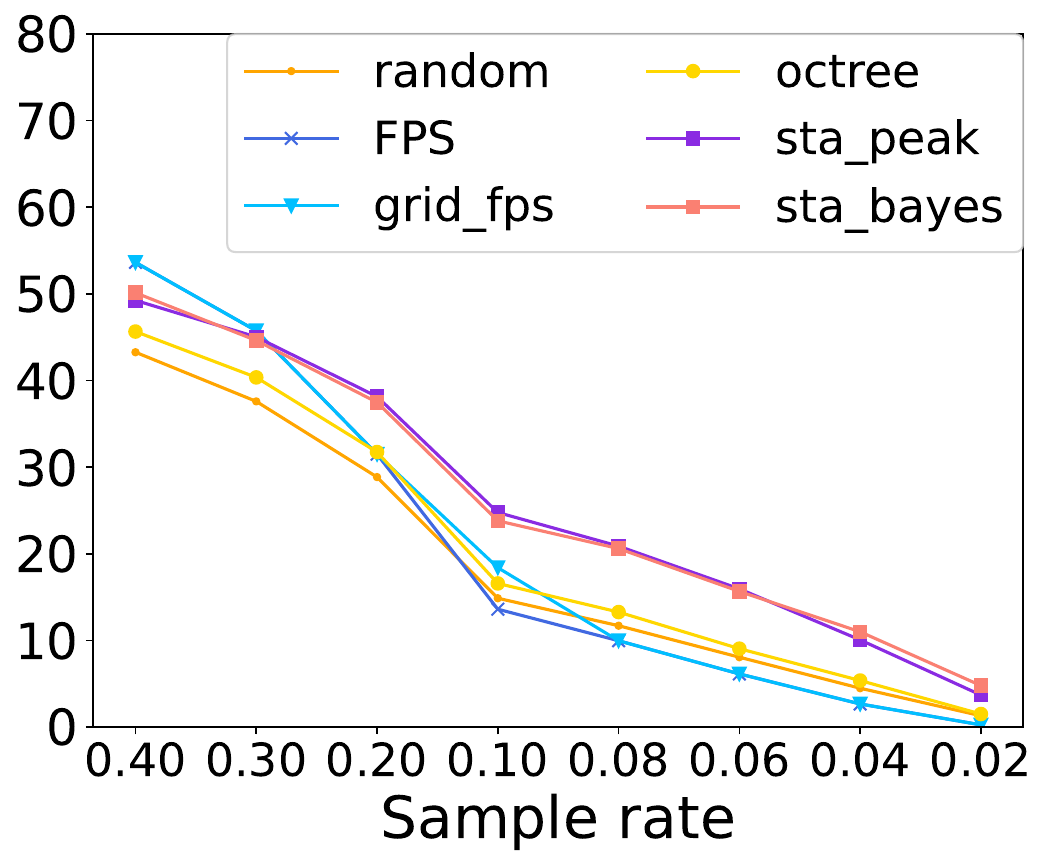}
        \caption{Hard (SECOND)}
        \label{fig5: f}
    \end{subfigure}
    \caption{Macro mAP with KITTI (SECOND)}
    \label{fig5_2}
\end{figure*}

\vspace*{-2ex}
\subsubsection{Evaluation Metrics}
\vspace*{-2ex}

\noindent\textbf{Effectiveness study.} We use the Mean Average Precision (mAP) to evaluate the extent our sampling method impacts the result of object detection~\cite{everingham2010pascal}.

\noindent\textbf{Efficiency study.} We measure the running time required to sample one point cloud.

\subsubsection{Parameter Settings}
\vspace*{-2ex}

\label{exp: setup_parameter}
\noindent\textbf{Determine the lower bound of the sample rate.} The resolution of point clouds varies across datasets. 
To ensure method accuracy, determining an appropriate sample rate becomes crucial. This value was derived through the statistics from the dataset, with 10\% for KITTI and 7\% for nuScenes. Please refer to ``Lower Bound of Sample Rate" in our technical report~\cite{mygit} for details.

\noindent\textbf{Default Configurations.} For all our demonstrated experiments, we use KITTI with Part-A2 as the inference model at sample rate 30\%.

\subsection{Experiment Results}

\subsubsection{Effectiveness Study (Q1)}

\label{exp: result_effectiveness}
To evaluate the effectiveness of the proposed statistical sampling method, we conducted experiments across a range of sampling rates, ranging from 30\% to 10\%. To have a comprehensive view of the methods, we analyzed categorical mAP (along with instance recall rate), and macro mAP. As introduced in Section \ref{exp: setup_parameter}, the sampling rate lower bound for KITTI and nuScenes was 10\% and 7\%, respectively. We demonstrated our experiments from 30\% to 10\% for KITTI, and from 10\% to 6\% for nuScenes. More experiment results with a wider range of sample rates and the corresponding evaluation result on Instance Recall Rate can be found in ``Results on More Sample Rates" and ``Study of Instance Recall Rate" of our technical report~\cite{mygit}.

\noindent\textbf{a) Comparison with Baselines under Categorical mAP.} 
For better demonstration, our analysis was conducted using Part-A2 on KITTI (Figure \ref{fig1}). Here are our observations: 

First, our proposed two methods, statistic\_peak (denoted as sta\_peak) and statistic\_bayes (denoted as sta\_bayes), both distinctively outperformed other baselines, except for FPS with a high sampling rate of 30\%, as shown in Figure \ref{fig1: a}. 
The advantage became more obvious in lower sampling rates (in Figure \ref{fig1: b}-\ref{fig1: c}).

Second, at a high sampling rate, i.e., 30\%, FPS showed exceptional performance, especially in categories `car' and `cyclist', as shown in Figure \ref{fig1: a}. FPS effectively preserved point diversity, rendering an excellent performance on a high sampling rate and on larger objects. However, the performance experienced a steep decline when the sampling rate decreased ($\leq$20\%) or with small objects, e.g. pedestrian (ped), as shown in Figure \ref{fig1: b}-\ref{fig1: c}. 

Third, grid\_fps, as a baseline we proposed, was designed to mitigate the drawback of FPS in both efficiency (which can be found in Section \ref{exp: result_efficiency}) and effectiveness at low sampling rates. With grid\_fps, we divided the point cloud into small grids and performed FPS in each grid. From Figure \ref{fig1}, we can observe that grid\_fps was a strong baseline and surpassed almost all others except sta\_peak and sta\_bayes.

Overall, our methods, sta\_peak and sta\_bayes had the advantage in both effectiveness
and robustness at various sample rates compared to baselines.

\begin{figure}[h]
    \centering
    \includegraphics[width=0.35\linewidth]{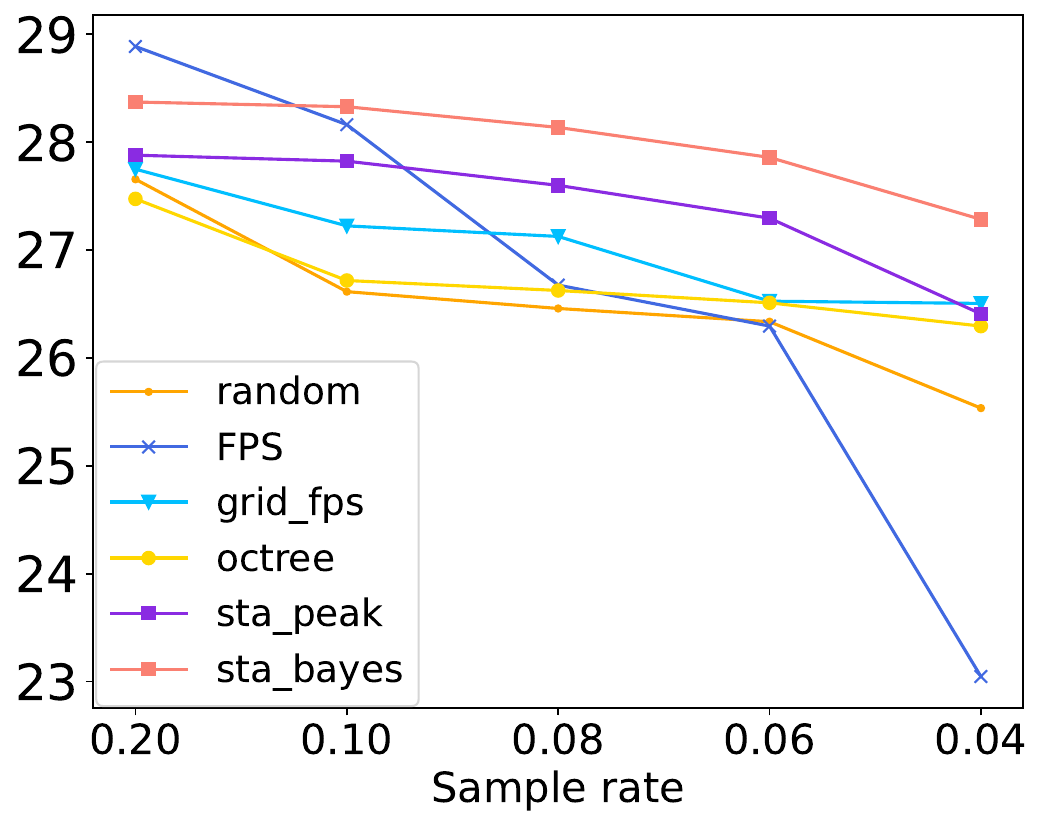}
    \caption{Macro mAP with PointPillars on nuScenes}
    \label{fig4: e}
    \vspace*{-2.5ex}
\end{figure}

\noindent\textbf{b) Comparison with Baselines under Macro mAP.} 
Figure \ref{fig5}
illustrated the macro mAP with PartA2 (KITTI). First, for all examined methods, a decline in sampling rate correlates with a decrease in mAP, and FPS demonstrated the most rapid decline. The mAP falls below that of random sampling when the sampling rate falls below 20\%. Second, sta\_peak and sta\_bayes showed a more significant advantage when the sampling rate decreased, indicating that they were capable of effectively downsampling the point cloud by maintaining a high mAP. Third, we assessed the macro mAP in accordance with different levels of hardness, which is defined based on the occlusion, and object size \cite{kitti2019iccv}. 

We found that sta\_peak and sta\_bayes consistently held a distinct advantage across all hardness levels, surpassing all other baselines. By contrast, FPS and grid\_FPS gradually lost their advantage at moderate and hard levels. 

\subsubsection{Density Peak vs. Na\"ive Bayes (Q2)}

\label{exp: peak_vs_bayes}

\noindent\textbf{a) Density Peak Method vs. Na\"ive Bayes Method.} 
Upon examining the categorical mAP, e.g., Figure \ref{fig1} for mAP with Part-A2 (KITTI), at a sampling rate of 30\% and 20\%, sta\_bayes could hardly outperform sta\_peak overwhelmingly, especially in the category `car'. However, the advantage of sta\_bayes gradually became apparent when the sampling rate decreased, especially for small objects (e.g., pedestrians). 
For the macro mAP, as shown in Figure \ref{fig5},
sta\_bayes consistently outperformed sta\_peak at all hardness levels. This indicates that sta\_bayes excelled in detecting objects with diverse truncations and visibilities, demonstrating enhanced effectiveness. 

\noindent\textbf{b) Generalization Ability.} In all the aforementioned comparisons, experiments were conducted across different datasets using various inference models, yielding consistent conclusions. 
Specifically, for the KITTI dataset, categorical mAP results were depicted in Figure~\ref{fig1} and Figure~\ref{fig3}, with inference models Part-A2 and SECOND, respectively. Additionally, Figures~\ref{fig5} and~\ref{fig5_2} present the macro mAP for these two models. Similarly, experiments were conducted on the nuScenes dataset using PointPillars as the inference model, with results illustrated in Figure~\ref{fig4} and Figure~\ref{fig4: e}.

\vspace*{-2.5ex}
\subsubsection{Efficiency Study (Q3)} 
\label{exp: result_efficiency}
\vspace*{-2ex}



\begin{table}[h]
  \centering
  \small
  \caption{Processing time per point cloud in seconds}
  \begin{tabular}{
    l
    S[table-format=1.4]
    S[table-format=1.2]
    S[table-format=1.2]
    S[table-format=1.2]
    S[table-format=1.2]
    S[table-format=1.2]
  }
    \toprule
    {Dataset} & {rand} & {FPS} & {g\_fps} & {octree} & {\textbf{sta\_p}} & {\textbf{sta\_b}} \\
    \midrule
    KITTI    & 0.0005 & 2.44 & 0.65 & 0.68 & 0.05 & 0.09 \\
    nuScenes & 0.001  & 8.13 & 2.47 & 1.36 & 0.13 & 0.32 \\
    \bottomrule
  \end{tabular}
  \label{table: time}
\end{table}
Table \ref{table: time} provided the time required by various methods for sampling a single point cloud. First, for our proposed two methods, sta\_peak showed an advantage compared with baselines, while sta\_bayes exhibited an obvious time increase compared with sta\_peak. According to our analysis in Section \ref{exp: peak_vs_bayes}, sta\_bayes traded the processing time for better sample results. 
Second, the KITTI dataset typically consists of approximately 18,000 points per point cloud, whereas nuScenes exhibited approximately 30,000 points per point cloud. Specifically, the LiDAR scanning rate for KITTI operated at 10Hz, resulting in a point cloud generation rate of 0.1 seconds per point cloud. In contrast, nuScenes adopted a data collection rate of 2Hz, equal to 0.5 seconds per point cloud. It is noteworthy that, among the methods we evaluated, only sta\_peak, sta\_bayes, along with the random sampling, had the capability to meet the generation rates for both datasets.

\subsubsection{Ablation Studies (Q4)}
\label{sec: exp_ablation}

\begin{figure}[h]
    \centering
    \includegraphics[scale=0.17]{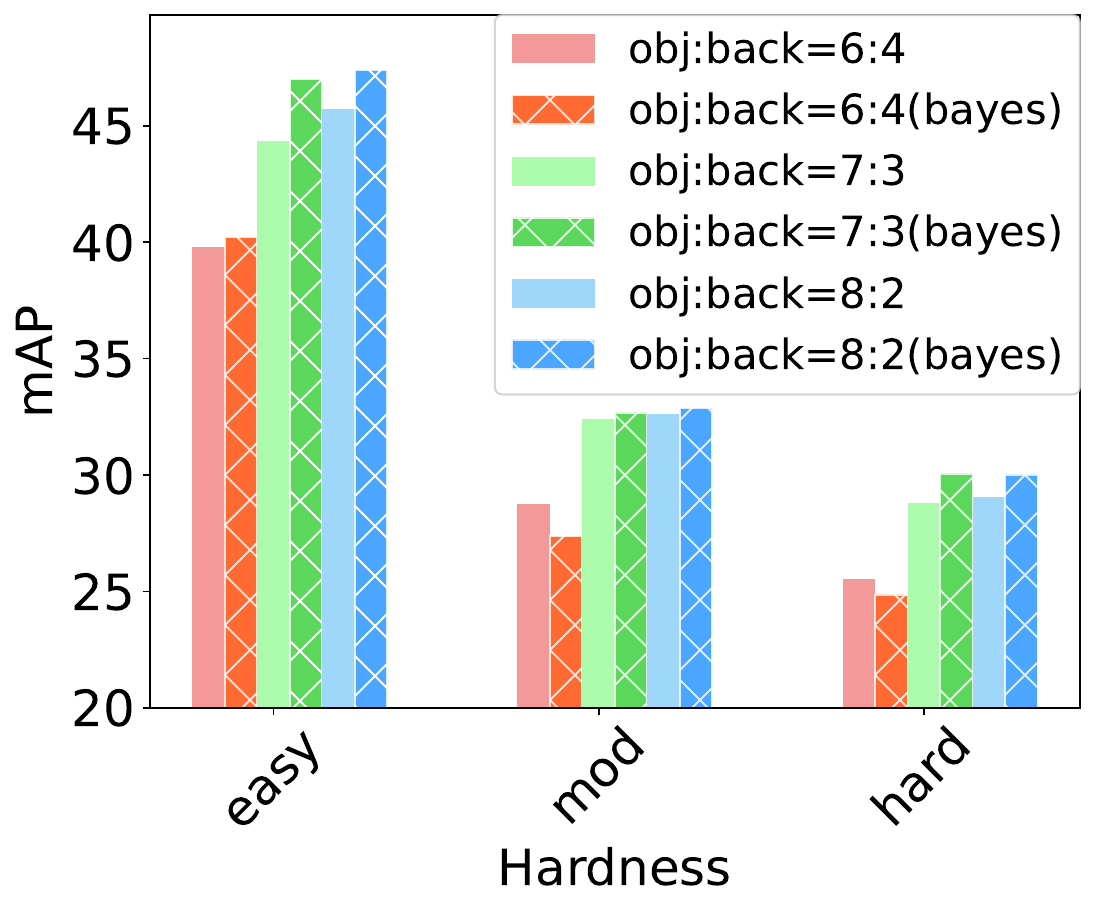}
    \caption{different object and background ratio}
    \label{fig: ablation_variant}
    \vspace*{-2ex}
\end{figure}

\noindent\textbf{Effect of Different Background Ratio.} We conducted experiments on different object-to-background ratios for sampled points using both sta\_peak and sta\_bayes, as depicted in Figure~\ref{fig: ablation_variant}.  We experimented with the ratios of 6:4, 7:3, and 8:2, respectively. The results showed that, the more object points we sampled, the higher mAP we obtained. However, there was a substantial mAP increase from the 6:4 to the 7:3, while the increase was less pronounced from the 7:3 to the 8:2, which suggested that increasing the ratio of object points can indeed enhance overall accuracy.

\noindent\textbf{Effect of Z-axis Filtering.}
Table~\ref{table: z-axis} presented the experiment results of performing the point classification with the Z-axis filter and without the Z-axis filter. The results indicated that the Z-axis filter provided more benefits at sample rates higher than 10\%, but the advantage gradually disappeared as the sample rate decreased. The experiment results indicated that the Z-axis filter could effectively filter out the ground, which significantly contributed to retaining object points. However, this method inevitably led to the issue of classifying the bottom part of objects on the ground as background points. Consequently, at low sample rates, using the Z-axis filter resulted in a lower mAP.

\begin{table}[h]
  \centering
  \small
  \caption{Ablation study on z-axis for density peak method}
  \begin{tabular}{
    S[table-format=2.0]%
    S[table-format=2.2]
    S[table-format=2.2]
    S[table-format=2.2]
    S[table-format=2.2]
    S[table-format=2.2]
    S[table-format=2.2]
  }
    \toprule
    {Rate} & \multicolumn{3}{c}{without z-axis filter} & \multicolumn{3}{c}{\textbf{with z-axis filter}} \\
    \cmidrule(lr){2-4} \cmidrule(lr){5-7}
    & {Easy} & {Med} & {Hard} & {Easy} & {Med} & {Hard} \\
    \midrule
    50\% & 74.26 & 60.49 & 56.96 & 75.39 & 61.01 & 57.35 \\
    30\% & 66.65 & 52.89 & 49.34 & 67.50 & 52.13 & 48.84 \\
    10\% & 44.33 & 32.41 & 28.81 & 42.19 & 29.89 & 26.65 \\
    6\%  & 29.50 & 20.92 & 17.89 & 29.75 & 19.91 & 17.23 \\
    2\%  &  8.23 &  5.15 &  4.10 &  6.05 &  3.72 &  3.16 \\
    \bottomrule
  \end{tabular}
  \label{table: z-axis}
\end{table}

\vspace*{-2ex}
\section{Conclusion}

The excessive amount of 3D points used in autonomous driving is a great challenge to processing efficiency and hardware cost. To tackle the problem, we introduce the Representative Point Selection approach, aiming at downsampling point clouds with awareness to object presence.
We propose a two-step sampling approach to efficiently downsample point clouds, focusing on retaining object points. Through experiments on KITTI and nuScenes, the method proves more effective and efficient than existing baselines.

\newpage
{\small
\bibliographystyle{ieee_fullname}
\bibliography{egbib}
}

\end{document}